\documentclass[journal]{IEEEtran}

\usepackage{graphicx,epsfig,subfigure,epstopdf}  
\usepackage{amsfonts,amssymb,amsmath,latexsym,amsthm}
\usepackage{bm}
\usepackage[ruled,linesnumbered]{algorithm2e}
\usepackage{algpseudocode}
\usepackage{multirow}
\usepackage{array}
\usepackage{float}
\usepackage{color}
\usepackage{url}
\usepackage{hyperref}
\usepackage[square,comma,numbers,sort&compress]{natbib}
\usepackage{booktabs}
\usepackage{enumitem}

\DeclareMathOperator*{\argmax}{arg\,max}

\DeclareMathOperator*{\dist}{dist}
\newcommand*{\dif}{\mathop{}\!\mathrm{d}}

\newtheorem{remark}{Remark}

\begin{document}
\title{Instance Weighted Incremental Evolution \\ Strategies for Reinforcement Learning \\ in Dynamic Environments}
\author{Zhi~Wang,~\IEEEmembership{Member,~IEEE},~Chunlin~Chen,~\IEEEmembership{Senior Member,~IEEE},~Daoyi~Dong,~\IEEEmembership{Senior Member,~IEEE}
\thanks{Manuscript accepted by \textit{IEEE Transactions on Neural Networks and Learning Systems}, 2022, DOI: 10.1109/TNNLS.2022.3160173.}
\thanks{The work is supported in part by the National Natural Science Foundation of China under Grant 62006111 and Grant 62073160; in part by the Australian Research Council's Discovery Projects funding scheme under Project DP190101566; in part by the Natural Science Foundation of Jiangsu Province of China under Grant BK20200330; and in part by the Alexander von Humboldt Foundation, Germany. (\textit{Corresponding authors: Chunlin Chen; Daoyi Dong})}
\thanks{Z. Wang is with the Department of Control and Systems Engineering, Nanjing University, Nanjing 210093, China, and with the School of Engineering and Information Technology, University of New South Wales, Canberra, ACT 2600, Australia (email: zhiwang@nju.edu.cn).}
\thanks{C. Chen is with the Department of Control and Systems Engineering, Nanjing University, Nanjing 210093, China (email: clchen@nju.edu.cn).}
\thanks{D. Dong is with the School of Engineering and Information Technology, University of New South Wales, Canberra, ACT 2600, Australia (email: daoyidong@gmail.com).}
}

\maketitle

\begin{abstract}
Evolution strategies (ES), as a family of black-box optimization algorithms, recently emerge as a scalable alternative to reinforcement learning (RL) approaches such as Q-learning or policy gradient, and are much faster when many central processing units (CPUs) are available due to better parallelization.
In this paper, we propose a systematic incremental learning method for ES in dynamic environments.
The goal is to adjust previously learned policy to a new one incrementally whenever the environment changes.
We incorporate an instance weighting mechanism with ES to facilitate its learning adaptation, while retaining scalability of ES.
During parameter updating, higher weights are assigned to instances that contain more new knowledge, thus encouraging the search distribution to move towards new promising areas of parameter space. 
We propose two easy-to-implement metrics to calculate the weights: instance novelty and instance quality.
Instance novelty measures an instance's difference from the previous optimum in the original environment, while instance quality corresponds to how well an instance performs in the new environment.
The resulting algorithm, Instance Weighted Incremental Evolution Strategies (IW-IES), is verified to achieve significantly improved performance on challenging RL tasks ranging from robot navigation to locomotion. 
This paper thus introduces a family of scalable ES algorithms for RL domains that enables rapid learning adaptation to dynamic environments.
\end{abstract}



\section{Introduction}
\IEEEPARstart{I}{n} reinforcement learning (RL)~\citep{sutton2017reinforcement}, an agent learns to perform a sequence of actions in an environment that maximizes cumulative reward based on the Markov decision process (MDP) formalism~\citep{xu2007kernel,luo2016model,yu2018reusable,li2019deep,li2020quantum}.
A primary driving force behind the explosion of RL is its integration with powerful nonlinear function approximators like deep neural networks (DNNs), aiming to develop agents that can accomplish challenging tasks in complex and uncertain environments.  
This partnership with deep learning, i.e., deep reinforcement learning (DRL), has enabled RL to successfully extend to tasks with high-dimensional state and action spaces, ranging from arcade games~\citep{mnih2015human}, board games~\citep{silver2017mastering} to robotic control tasks~\citep{duan2016benchmarking}.

An alternative approach to solving RL problems is using black-box optimization, known as direct policy search~\citep{heidrich2009hoeffding} or neuroevolution~\citep{risi2017neuroevolution} when applied to neural networks.
Evolution strategies (ES)~\citep{beyer2013theory} are a particular family of these optimization algorithms that are heuristic search procedures inspired by natural evolution.
Recent research has reported that ES can be competitive to popular backpropagation-based algorithms such as policy gradient and Q-learning on challenging RL problems, with much faster training speed when many central processing units (CPUs) are available due to better parallelization~\citep{salimans2017evolution}.
ES can reliably train neural network policies, in a fashion well suited to scale up to modern distributed computer systems without requirements for temporal discounting, backpropagating gradients and value function approximation~\citep{conti2018improving,khadka2018evolution,liu2019trust}.
The promising properties of applying ES for solving RL problems include:
\begin{enumerate}
\item Since ES only needs to communicate scalar returns of complete episodes, it is highly parallelizable and enables near-linear speedups in runtime as a function of CPUs.
\item ES uses a fitness metric that consolidates returns across an entire episode, making it invariant to sparse or deceptive rewards with arbitrarily long time horizons.
\item The population-based evolutionary search provides diverse exploration, particularly when combined with explicit diversity maintenance techniques.
Moreover, the redundancy inherent in a population also facilitates robustness and stable convergence properties, especially when incorporated with elitism.
\end{enumerate}

Traditional research on ES algorithms for RL tasks mainly focuses on stationary optimization problems, which are precisely given in advance and remain fixed during the entire evolutionary process.
Instead, the environments of real-world RL applications are often dynamic, where the state space, available actions, state transition functions or reward functions may change over time instead of being static, such as for multi-agent cases~\citep{zhou2017multiagent}, robot navigation problems~\citep{jaradat2011reinforcement} or online learning settings~\citep{nagabandi2019learning}.
This challenge leads to the dynamic optimization problems~\citep{jiang2018transfer,yang2008population} for corresponding ES algorithms where the fitness function, design variables, or environmental conditions change over time.

In the paper, we tackle the dynamic environment as a sequence of stationary tasks on a certain timescale where each task corresponds to a stationary environment during the associated time period.
Learning in such dynamic environments is characterized not only by the capability of acquiring complex skills, but also the ability to adapt rapidly under a non-stationary task distribution.
Humans and animals can learn complex models that precisely and reliably reason about real-world phenomena, and they can rapidly adjust such models in the face of unexpected changes.
Although (deep) neural network models can represent very complex functions, they lack the capability of rapidly adapting to dynamic environments.
To circumvent the necessity for repeatedly re-evolving, recent research exploits transfer learning techniques~\citep{pan2018multisource} as a tool to reuse information available from a set of source tasks to help the evolutionary performance in a related but different target task~\citep{jiang2018transfer}.
Generally, it requires repeatedly accessing and processing a potentially large distribution of source tasks to provide a good knowledge base for target environments that are supposed to be consistent with the source distribution.

An increasing number of real-world scenarios requires RL algorithms to be capable of adapting their behaviors \textit{in an incremental manner} to environments that may drift or change from their nominal situations, continuously utilizing previous knowledge to benefit the future decision-making process.
Hence, incremental learning~\citep{wang2019tmechl,wang2019tnnls,wang2021lifelong} emerges by incrementally adjusting the previously learned policy to a new one whenever the environment changes,
\footnote{In this setting, the policy parameters of the new environment are initialized from the previously learned optima of the original environment.
The reason is that, the previous optimal policy empirically performs better than a randomly initialized one, since it has learned some of the features (e.g., nodes in the neural network) of the state-action space.
This procedure is akin to the pre-training in the deep learning community, where layers in a neural network extract hierarchical levels of feature representation. 
Model parameters pre-trained on common datasets, such as ImageNet~\citep{russakovsky2015imagenet}, can be used as a helpful initialization for general downstream tasks~\citep{yosinski2014transferable}.}
which offers an appealing alternative that is amenable for rapid learning adaptation to dynamic environments.
Such an incremental adaptation is crucial for intelligent systems operating in the real world, where changing factors and unexpected perturbations are the norm. 
Incremental learning has been widely investigated to cope with learning tasks with an ever-changing environment~\citep{he2011incremental}, in areas such as supervised learning~\citep{elwell2011incremental}, RL~\citep{wang2019tmechl,wang2019tnnls}, machine vision~\citep{ross2008incremental}, human-robot interaction~\citep{kulic2012incremental}, and system modeling~\citep{wang2018incremental}.
However, an equivalent notion of incremental learning in ES for RL domains has largely eluded researchers, with few related work available in the literature.
Here, we aim to develop a new incremental learning framework for the derivative-free ES algorithms, which is orthogonal and complementary to the previous one in~\citep{wang2019tnnls} that is investigated for the derivative-based RL approaches.



In the paper, we formulate an incremental learning procedure that uses natural evolution strategies (NES) to update the parameters of a policy network for RL in dynamic environments.
To increase the capability of rapid learning adaptation, we incorporate an \emph{instance weighting} mechanism with ES to improve the learning adaptation, while not sacrificing the speed/scalability benefits of ES.
During parameter updating, we assign higher weights to instances that contain more knowledge on the new environment, thus encouraging the search distribution to move towards new promising areas in the parameter space.
We propose two easy-to-implement metrics for calculating the weights: \emph{instance novelty} and \emph{instance quality}.
First, instance novelty intends to indicate the instance's difference from the previous optimum in the original environment, with the help of a domain-dependent behavior characterization that describes the behavior of the associated policy.
Second, instance quality corresponds to how well the instance performs in the new environment, where its performance is evaluated by the received return of the associated policy.
Together, instances with high weights are supposed either to differ more from the original environment (high novelty) or to be more in line with the new environment (high quality).
The resulting algorithm, Instance Weighted Incremental Evolution Strategies (IW-IES), ``reinforces" the evolutionary process of searching for well-behaving policies that fit in the new environment, thus facilitating more rapid learning adaptation to dynamic environments.

We test whether IW-IES improves the performance of ES on challenging RL tasks ranging from robot navigation to locomotion in dynamic environments. 
Experimental results confirm that IW-IES is capable of handling various dynamic environments, and achieves significantly rapid learning adaptation to these tasks.
In summary, the main contributions are listed as follows.
\begin{enumerate}
\item We introduce Instance Weighted Incremental Evolution Strategies (IW-IES), a family of scalable ES algorithms that addresses challenging RL problems in dynamic environments from an incremental learning perspective.
\item We incorporate an instance weighting mechanism with ES to facilitate learning adaptation to dynamic environments, while retaining scalability benefits and enabling a near-linear speedup in runtime as more CPUs are used. 
\item We propose two easy-to-implement metrics for calculating the weights: instance novelty and instance quality, which effectively enhance the evolutionary performance almost without extra computational complexity.
\item We perform extensive experiments to verify that IW-IES can consistently improve learning adaptation to dynamic environments over various state-of-the-art baselines.
\end{enumerate}

The rest of this paper sequentially presents the background on ES algorithms for RL domains in Section II, the proposed algorithm with designed weighting metrics in Section III, the experiments in Section IV, and the conclusions in Section V.

\section{Background}
\subsection{Evolution Strategies for Reinforcement Learning}
Reinforcement learning (RL) is commonly studied based on the Markov decision process (MDP) formalism.
An MDP is a tuple $\langle S,A,T,R,\gamma\rangle$, where $S$ is the set of states, $A$ is the set of actions, $T:S\times A\times S\to [0,1]$ is the state transition probability, $R:S\times A\to\mathbb{R}$ is the reward function, and $\gamma \in (0,1]$ is the discounting factor.
A policy is defined as a function $\pi:S\times A\to [0,1]$, a probability distribution that maps actions to states, and $\sum_{a\in A}\pi(a|s)=1,\forall s\in S$.
The goal of RL is to find an optimal policy $\pi^*$ that maximizes the expected long-term return $J(\pi)$:
\begin{equation}
J(\pi) = \mathbb{E}_{\tau\sim \pi(\tau)}[r(\tau)] =  \mathbb{E}_{\tau\sim \pi(\tau)}\left[\sum\nolimits_{i=0}^{\infty}\gamma^ir_i\right],
\label{Jpi}
\end{equation}
where $\tau=(s_0, a_0, s_1, a_1, ...)$ is the learning episode, $\pi(\tau)=p(s_0)\Pi_{i=0}^{\infty}\pi(a_i|s_i)p(s_{i+1}|s_i,a_i)$, $r_i$ is the instant reward received when executing action $a_i$ in state $s_i$.

Inspired by natural evolution, ES is designed to cope with high-dimensional continuous-valued domains and has remained an active field of research for more than four decades~\citep{beyer2013theory}.
ES algorithms address the following search problem: maximize a non-linear fitness function that is a mapping from search space $\mathcal{S}\subseteq\mathbb{R}^d$ to $\mathbb{R}$.
At each iteration (generation), a population of parameter vectors (gnomes) is perturbed (mutated) and optionally recombined (merged) via crossover.
The mutation is usually carried out by adding a realization of a normally distributed random vector.
Each resultant offspring is evaluated by a fitness function, and the highest scoring parameter vectors are then recombined to form the population for the next generation.
Recent research highlights the scalability of ES algorithms on many high-dimensional RL tasks while offering unique benefits over traditional gradient-based RL methods~\citep{conti2018improving}.
Most notably, ES is highly parallelizable and well suited to modern distributed computer systems with a near-linear speedup in wall-clock runtime.
\citet{salimans2017evolution} reported that, with hundreds of parallel CPUs, ES is able to achieve roughly the same performance on Atari games with the same DNN architecture in 1 hour as A3C~\citep{mnih2016asynchronous} did in 24 hours.

Algorithms in the ES class differ in their representations of population and methods of recombination.
The version of ES used in this paper belongs to the class of natural evolution strategies (NES)~\citep{wierstra2014natural}, which constitutes a well-principled approach with a clean derivation from first principles.
The core idea is to iteratively update parameters of the search distribution using the sampled gradient of expected fitness. 
The search distribution can be taken to be a multinormal distribution, but could in principle be any distribution of which the log-density is differentiable.
Let $\bm{\theta}$ denote parameters of the search distribution's density $p(\bm{z}|\bm{\theta})$, and $f(\bm{z})$ denotes the fitness function (e.g., received return) for instance $\bm{z}$.
At each generation, a population of search instances is produced by the parameterized search distribution, and the fitness function is evaluated at each instance.
The expected fitness under the search distribution is written as:
\begin{equation}
J(\bm{\theta}) = \mathbb{E}_{\bm{\theta}}[f(\bm{z})] = \int f(\bm{z})p(\bm{z}|\bm{\theta})\dif\bm{z}.
\label{objective}
\end{equation}
In a fashion similar to REINFORCE~\citep{williams1992simple}, NES takes gradient steps on $\bm{\theta}$ with the following estimator:
\begin{equation}
\begin{aligned}
\nabla_{\bm{\theta}}J(\bm{\theta}) & = \nabla_{\bm{\theta}}\int f(\bm{z})p(\bm{z}|\bm{\theta})\dif\bm{z} \\
& = \mathbb{E}_{\bm{\theta}}[f(\bm{z})\nabla_{\bm{\theta}}\log p(\bm{z}|\bm{\theta})].
\end{aligned}
\end{equation}
We can obtain the Monte Carlo estimate of the search gradient for instances in a population ($\bm{z}_1,...,\bm{z}_m$) as:
\begin{equation}
\nabla_{\bm{\theta}}J(\bm{\theta}) \approx \frac{1}{m}\sum_{i=1}^mf(\bm{z}_i)\nabla_{\bm{\theta}}\log p(\bm{z}_i|\bm{\theta}),
\end{equation}
where $m$ is the population size. 
For each generation, NES estimates a search gradient on the parameters towards higher expected fitness in promising regions. 
Instead of using the plain stochastic gradient for updates, NES follows the natural gradient, which helps mitigate the slow convergence of plain gradient ascent in optimization landscapes with ridges and plateaus. 
The direction of the natural gradient is associated with the \emph{Fisher information matrix} of the given parametric family of the search distribution:
\begin{equation}
\begin{aligned}
\bm{F} & = \int p(\bm{z}|\bm{\theta})\nabla_{\bm{\theta}}\log p(\bm{z}|\bm{\theta})\nabla_{\bm{\theta}}\log p(\bm{z}|\bm{\theta})^T\dif\bm{z}  \\
& = \mathbb{E}[\nabla_{\bm{\theta}}\log p(\bm{z}|\bm{\theta})\nabla_{\bm{\theta}}\log p(\bm{z}|\bm{\theta})^T].
\end{aligned}
\end{equation}
If $\bm{F}$ is invertible, the natural gradient amounts to 
\begin{equation}
\widetilde{\nabla}_{\bm{\theta}}J(\bm{\theta}) = \bm{F}^{-1}\nabla_{\bm{\theta}}J(\bm{\theta}).
\end{equation}
The local structure of the fitness function is adaptively captured by the search distribution's parameters, e.g., the mean and covariance matrix in a Gaussian distribution.
The evolutionary process re-iterates until a stopping criterion is met.

In RL domains, NES directly searches in the parameter space of a neural network to find an effective policy. 
For scalability to high-dimensional problems, the population $\{\bm{z}_i\}_{i=1}^m$ is typically instantiated as a multivariate Gaussian with diagonal covariance matrix centered at $\bm{\theta}$, i.e., $\bm{z}_i=\bm{\theta}+\sigma\bm{\epsilon}_i$, where $\sigma$ is the noise standard deviation. 
The following gradient estimator:
\begin{equation}
\nabla_{\bm{\theta}}\mathbb{E}_{\bm{\epsilon}\sim \mathcal{N}(\bm{0},\bm{I})}[f(\bm{\theta}+\sigma\bm{\epsilon})] = \frac{1}{\sigma}\mathbb{E}_{\bm{\epsilon}\sim \mathcal{N}(\bm{0},\bm{I})}[f(\bm{\theta}+\sigma\bm{\epsilon})\bm{\epsilon}]
\label{gradient}
\end{equation}
can be estimated with samples:
\begin{equation}
\nabla_{\bm{\theta}}f(\bm{\theta}) \approx \frac{1}{m\sigma}\sum_{i=1}^{m}f(\bm{\theta}+\sigma\bm{\epsilon}_i)\bm{\epsilon}_i,
\end{equation}
and then parameters $\bm{\theta}$ are updated iteratively by $\bm{\theta}\leftarrow\bm{\theta}+\alpha\nabla_{\bm{\theta}}J(\bm{\theta})$ till convergence, where $\alpha$ is the learning rate.
In this way, the gradient estimation reduces to sampling unit Gaussian perturbation vectors $\bm{\epsilon}\!\sim\!\mathcal{N}(\bm{0},\bm{I})$, evaluating the performance (fitness) of the perturbed policies and aggregating the results over a population of search instances.

If the random seeds between workers are synchronized before optimization, each worker can know the perturbations used by other workers.
In this way, only a single scalar (fitness) needs to be communicated among the workers to agree on a parameter update, thus resulting in highly parallelizable implementations.
More details about ES can be found in~\citep{wierstra2014natural,salimans2017evolution}.

\subsection{Related Work}
While RL algorithms have demonstrated the ability to learn control policies for complex and high-dimensional problems, it is still challenging to apply them to tasks in dynamic environments.
A related class of methods in the context of dynamic environments is transfer reinforcement learning~\citep{pan2018multisource}, which reuses the knowledge from a set of related source domains to help the learning task in the target domain.
One feasible approach is to use \textit{domain randomization} to learn policies that can work under a large variety of environments.
Training a robust policy with domain randomization has been shown to improve the transfer from simulation to reality, also known as ``sim-to-real".
\citet{tobin2017domain} first trained an object detector with randomized appearances in simulation  and transferred it to perform the real robotic grasping task. 
\citet{muratore2019assessing} randomized the parameters of the physics simulations to train robust policies that can be applied directly to real second-order nonlinear systems with an approximate probabilistic guarantee on the suboptimality.
\citet{sheckells2019using} showed that, with the aid of fitting a stochastic dynamics model, the learned robust policy can be transferred back to the real vehicle with little loss in predicted performance.
These methods require task-specific knowledge to design parameters and the range of the randomized domain.
The policy trained over an enormous range of domains may learn a conservative strategy or fail to learn the target task, while a small range may be insufficient of providing sufficient variation for the policy to transfer to uncertain environments.

Instead of learning invariance to environmental dynamics, an alternative solution is to formulate an environment-conditioned policy as a function of the current state and task feature.
\citet{chen2018hardware} proposed an explicit representation of the hardware variations and used it as additional input to the policy function for each discrete instance of the environment. 
\citet{yu2017preparing} incorporated an online system identification module with history observations to explicitly predict the dynamics parameters, which are provided as the input to a policy to compute appropriate controls.
Subsequently, \citet{yu2019policy} leveraged domain randomization for learning a family of policies conditioned on explicit environmental dynamics. 
When tested in unknown environments, it directly searched for the best policy in the family based on the task performance via covariance matrix adaptation evolution strategies (CMA-ES).
Constructing these environment-conditioned policies necessarily requires structural assumptions about the system's dynamics, which may not hold in the real world. 
In addition, it may be difficult for more complex systems to identify the dynamics parameters at runtime.

A similar idea is to train an adaptive policy that is able to identify the environmental dynamics and apply actions appropriate for different system dynamics.
In the absence of direct knowledge of parameters of interest, the dynamics can be inferred from a history of past states and actions.
\citet{peng2018sim} implicitly embedded the system identification module into the policy by using a recurrent model, where the internal memory acts as the summary of past states and actions, thereby providing a mechanism for inferring the system's dynamics from the policy itself.
\citet{andrychowicz2020learning} formulated memory-augmented recurrent polices for in-hand manipulation tasks, which admits the possibility to learn an adaptive behavior and implicit system identification on the fly. 
These adaptive policies can be trained in the assumed source tasks and deployed in the unknown dynamic environment without fine-tuning.
However, policies trained over the source distribution may not generalize well when the discrepancy between the target environment and the source is too large.

Another line of research that tries to address dynamic environments is meta-learning, also called learning-to-learn~\citep{lake2017building}.
A recent trend in meta-learning is to find good initial weights from which adaptation can be quickly performed to tasks sampled from a distribution.
One such approach is the gradient-based model-agnostic meta-learning (MAML) algorithm~\citep{finn2017model}.
\citet{gajewski2019evolvability} derived a novel objective that maximizes the diversity of exhibited behaviors and explicitly optimizes the \textit{evolvability} of ES algorithms, i.e., the ability to further adapt to changing circumstances.
\citet{houthooft2018evolved} evolved a differentiable loss function that is meta-trained via temporal convolutions over the agent's experiences, resulting in faster test time learning on novel tasks sampled from the same distribution.
\citet{song2020esmaml} employed ES algorithms to solve the MAML problem and to train the meta-policy without estimating any second derivatives.
In general, existing methods require repeatedly accessing and processing a potentially large distribution of source tasks to provide a reliable knowledge base for target environments that are supposed to be consistent with the source distribution.
In contrast, our incremental learning mechanism concentrates on the ability to rapidly learn and adapt in a sequential manner, without any structural assumptions or prior knowledge on the dynamics of the ever-changing environment.

\section{Instance Weighted Incremental Evolution Strategies (IW-IES)}
In this section, we first formulate the incremental learning procedure to address ES algorithms in a dynamic environment.
Then, we present the framework of incremental evolution strategies incorporated with the instance weighting mechanism.
Next, we introduce two easy-to-implement metrics of instance novelty and instance quality as well as their mixing variant to calculate the weights.
Finally, we give the integrated IW-IES algorithm based on the above implementations.

\subsection{Problem Formulation}
Throughout the paper, we tackle the dynamic environment as a sequence of stationary tasks on a certain timescale. 
Each task corresponds to the specific environment characteristics during the associated time period.
The dynamic environment involves an infinite task distribution $\mathcal{D}$ over time:
\begin{equation}
\mathcal{D} = [M_1,...,M_{t-1},M_t,...],
\label{td}
\end{equation}
where each $M_t\in\mathcal{M}$ denotes the specific MDP that is stationary during the $t$-th time period, and $\mathcal{M}$ denotes the space of MDPs.
We assume that the environment changes in terms of the reward and transition functions only, while keeping the same state-action space.
Suppose that in the ($t-1$)-th time period, the optimal parameters  $\bm{\theta}_{t-1}^*$ are obtained by evolving the search distribution as:
\begin{equation}
\bm{\theta}_{t-1}^* = \argmax_{\bm{\theta}\in\mathbb{R}^d} J_{M_{t-1}}(\bm{\theta}).
\end{equation}
When the environment changes to $M_t$, the goal of incremental learning is to adjust the previous optimum of policy parameters $\bm{\theta}_{t-1}^*$ to new $\bm{\theta}_t^*$ that fit in the new environment:
\begin{equation}
\bm{\theta}^*_t = \argmax_{\bm{\theta}\in\mathbb{R}^d}J_{M_t}(\bm{\theta}),
\end{equation}
with initialization of $\bm{\theta}_{t}\leftarrow\bm{\theta}^*_{t-1}$.
Continually, the optimum of policy parameters is incrementally adjusted to a new one, ($\bm{\theta}_{t+1}^*,\bm{\theta}_{t+2}^*,...$), whenever the environment changes.

\begin{remark}
As a matter of fact, automatic detection and identification of changes is also an important component of learning in dynamic environments. 
In the paper, we merely concentrate on how ES algorithms enable rapid learning adaptation to the new environment once the change has been detected and identified. 
It is analogous to investigations in fault tolerant control or dynamic multiobjective optimization~\citep{jiang2018transfer}, which solely focus on how controllers/algorithms can quickly accommodate dynamic changes while leaving the detection and identification to be approached individually.
\end{remark}

\subsection{Framework}
In the incremental learning setting, initializing policy parameters from the original environment empirically benefits the evolutionary process when starting to interact with the new environment, since the previous optimum has learned some of the feature representations of the state-action space.
However, the previous optimum of policy parameters may be a local one that has been overfitted to the original environment, especially when using a nonlinear function approximator such as the (deep) neural networks.
This potential drawback in the incremental initialization may degrade the performance of the new evolutionary process in the long term.

\begin{figure}[tb]
\centering
\subfigure[]{\includegraphics[width=0.48\columnwidth]{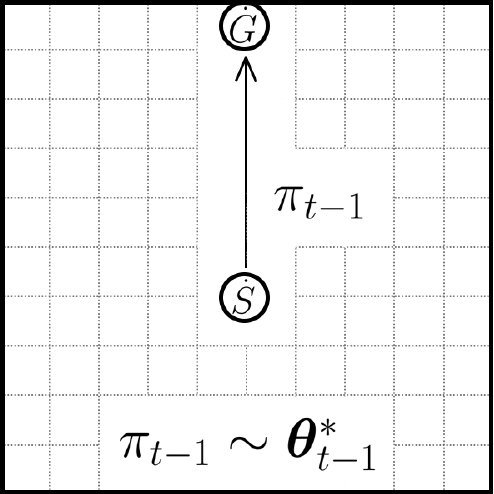}}
\subfigure[]{\includegraphics[width=0.48\columnwidth]{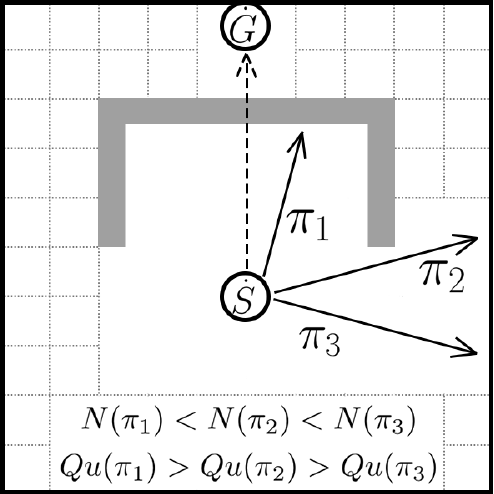}}
\caption{A simple example of the 2D navigation task in a dynamic environment. $\dot{S}, \dot{G}$ are the start and  goal points, and the black U-shaped object is a three-sided wall. (a) Original environment $M_{t-1}$. (b) New environment $M_t$, where $N(\cdot)$ denotes instance novelty and $Qu(\cdot)$ denotes instance quality.}
\label{fig:metircs}
\end{figure}

Unlike in supervised learning with DNNs, wherein local optima are not thought to be a problem~\citep{kawaguchi2016deep}, the training data in RL is determined by the executed policies associated with the search distribution.
Fig.~\ref{fig:metircs} illustrates a simple example of the 2D navigation task, where a three-sided wall blocks the previous optimal path to the goal in the new environment.
Due to not having adapted to the new environment yet, the search distribution tends to induce policies that perform well in the original environment (e.g., $\pi_1$ in Fig.~\ref{fig:metircs}-(b)), and move around regions in the parameter space adjacent to the previous optimum.
Therefore, the training data for the new evolutionary process is probably limited and it may not discover alternative strategies with potentially larger payoffs (e.g., $\pi_2$ or $\pi_3$).
Thus, it probably gets stuck in bad local optima.
Here we give another example.
Suppose that a navigation robot has learned how to reach a goal in the south direction.
When the goal changes to the north, the robot still tends to head south before it can slowly adapt to the new environment.

It can be inferred that, directly updating policy parameters from the previous optima probably hinders the search distribution to effectively explore the new environment, thus slowing down the learning adaptation.
To alleviate this problem, we incorporate an \emph{instance weighting}  mechanism with ES to improve its learning adaptation, while preserving the speed/scalability benefits of ES.
The idea is straightforward: during parameter updating, we assign higher weights to instances out of a population that contain more knowledge on the new environment, thus encouraging the search distribution to move towards new promising regions in the parameter space.

Recall the gradient estimator in (\ref{gradient}) where NES estimates the gradient by taking a sum of sampled parameter perturbations $\bm{\epsilon}$ weighted by their fitness $f(\bm{\theta}+\sigma\bm{\epsilon})$, where $\sigma$ is the noise standard deviation.
It rewards instances with high fitness, and encourages the search distribution to move towards the direction of those ``promising" instances.
In a similar spirit, we re-arrange the canonical objective function in (\ref{objective}) by multiplying it with a weight $w(\bm{z})$ assigned to each instance $\bm{z}$ as:
\begin{equation}
J(\bm{\theta}) = \mathbb{E}_{\bm{\theta}}[w(\bm{z})f(\bm{z})] = \int w(\bm{z})f(\bm{z})p(\bm{z}|\bm{\theta})\dif\bm{z}.
\end{equation}
Consequently, the gradient estimator in (\ref{gradient}) becomes:
\begin{equation}
\nabla_{\bm{\theta}}f(\bm{\theta}) \approx \frac{1}{m\sigma}\sum_{i=1}^{m}w(\bm{\theta}+\sigma\bm{\epsilon}_i)f(\bm{\theta}+\sigma\bm{\epsilon}_i)\bm{\epsilon}_i.
\end{equation}
Intuitively, the algorithm follows the approximated gradient in the parameter space towards instances that achieve high fitness of $f(\bm{\theta}+\sigma\bm{\epsilon}_i)$ and exhibit high weights of $w(\bm{\theta}+\sigma\bm{\epsilon}_i)$.
If the weighting metric can correctly indicate the amount of new knowledge contained by the instance, then the gradient estimator will reward instances with more new knowledge and encourages the search distribution to move towards new promising regions of parameter space that fit in the new environment.
The instance weighting mechanism ``reinforces" the evolutionary process that searches for well-behaving policies in the new environment, thus improving the learning adaptation to dynamic environments.

Based on the above insight, Algorithm~\ref{framework} presents the framework of the formalized incremental learning procedure.
It is clear that the performance highly depends on how the weight is calculated for each instance.
Next, we will introduce the weighting metrics of instance novelty, instance quality, and their mixing variant.

\begin{algorithm}[tb]
\caption{Incremental Learning Framework} \label{framework}
\KwIn{Current time period $t (t\ge 2)$; \newline 
learning rate $\alpha$; population size $m$; \newline 
noise standard deviation $\sigma$}
\KwOut{Optimal policy parameters $\bm{\theta}_t^*$ for $M_t$}
Initialize $\bm{\theta}_t \leftarrow \bm{\theta}_{t-1}^*$, and $m$ CPU workers with known random seeds \\
\While{not converged}{
	\For{\text{each worker} $i=1,...,m$}{
		Sample $\bm{\epsilon}_i\sim\mathcal{N}(\bm{0}, \bm{I})$ \\
		Compute fitness $f_i=f(\bm{\theta}_t+\sigma\bm{\epsilon}_i)$ \\
		Calculate and normalize instance weights $w_i=w(\bm{\theta}_t+\sigma\bm{\epsilon}_i)$ by metrics in Section III-C
	}
    Send all scalar fitness $f_i$ and weights $w_i$ from each worker to every other worker \\
    \For{\text{each worker} $i=1,...,m$}{
    	Reconstruct all perturbations $\bm{\epsilon}_i$ using known random seeds \\
    	Set $\bm{\theta}_t \leftarrow \bm{\theta}_t + \alpha\frac{1}{m\sigma}\sum_{i=1}^mw_if_i\bm{\epsilon}_i$
    }
}
\end{algorithm}

\subsection{Weighting Metrics}
\subsubsection{Instance Novelty}
Instance novelty, as our first metric, is designed to indicate the instance's difference from the previous optimum in the original environment.
Instances with high novelty are supposed to induce different policies from those performing well in the original environment, and hence probably contain more knowledge on the new environment.
As shown in Fig.~\ref{fig:metircs}-(b), compared to $\pi_1$, the example policy $\pi_2$ exhibits the behavior that differs more from the previous optimum, and is supposed to reveal higher instance novelty.

To attain feasible computation of such difference, one needs to hand-design or learn an abstract, holistic description of an agent's lifetime of behavior policy.
Let $\pi_{\bm{z}}$ denote the executing policy induced by instance $\bm{z}$. The policy is assigned a domain-dependent behavior characterization $b(\pi_{\bm{z}})$ that describes its behavior.
For example, in the case of a humanoid locomotion problem, $b(\pi_{\bm{z}})$ may be as simple as a two-dimensional vector containing the humanoid's final coordinate, or a concatenation of coordinates that records the humanoid's movement trajectory.
Throughout training in the original environment $M_{t-1}$, the final behavior characterization $b(\pi_{\bm{\theta}_{t-1}^*})$ can be obtained corresponding to the optimal instance $\bm{\theta}_{t-1}^*$.
Next, in the new environment $M_t$, a particular instance's novelty $N(\pi_{\bm{z}})$ is calculated by computing the distance between behavior characterizations of this instance and the previous optimum:
\begin{eqnarray}
N(\pi_{\bm{z}}) \hspace{-1.5em} && = \dist(b(\pi_{\bm{z}}),b(\pi_{\bm{\theta}_{t-1}^*})) \nonumber \\
&& = ||b(\pi_{\bm{z}}) - b(\pi_{\bm{\theta}_{t-1}^*})||_2~.
\end{eqnarray}
Here, the Euclidean distance ($L2$-norm) is used for behavior characterizations.
However, any distance function can be employed in principle.

Now, the calculated novelty is used as the first metric to assign the instance weight out of a population as:
\begin{equation}
w(\bm{z}_i) = m\cdot\frac{e^{ N(\pi_{\bm{z}_i})/\rho}}{\sum_{j=1}^me^{ N(\pi_{\bm{z}_j})/\rho}},~~i=1,...,m
\label{metric1}
\end{equation}
where $\rho\in\mathbb{R},\rho>0$ is the temperature hyperparameter for controlling the weight distribution.
When $\rho$ becomes larger, all $w'$s will be close to $1$, and this weighting metric reduces to uniform weighting.
In practice, we increase the temperature $\rho$ by a small increment $\Delta\rho$ at each parameter updating iteration.
As the evolution proceeds, the effect of instance weighting gradually becomes weak, and the exploration of novel behaviors decreases.
This procedure is akin to the classical ``exploration-exploitation trade-off" in the RL or evolutionary computation community, where exploration is progressively replaced by exploitation as the learning/evolution proceeds.

\begin{remark}
This metric is related to the concept of curiosity and seeking novelty in RL research and developmental robotics~\citep{oudeyer2007intrinsic}, which pushes a learning robot towards novel or curious situations.
The notion of novelty is also analogous to that of novelty search algorithms in the evolutionary computation community~\citep{lehman2011abandoning,conti2018improving}, which is inspired by nature's drive towards diversity and stimulates policies to explore different behaviors from those previously performed.
\end{remark}

\subsubsection{Instance Quality}
As it literally means, instance quality corresponds to how well the instance performs in the new environment.
Naturally, the performance is evaluated by the received return of the induced learning policy.
Since adjusting the previous optimum to a new one under a new data distribution could get stuck in bad local basins, assigning greater importance to high-quality instances can encourage the policies to move toward regions of parameter space that better fit in the new environment, which may be far away from the previous optimum.
As shown in Fig.~\ref{fig:metircs}-(b), due to environmental change, the two example policies, $\pi_2$ and $\pi_3$, cannot obtain a satisfactory learning performance yet in the new environment.
On the other hand, compared with $\pi_3$, the policy $\pi_2$ is more prone to inducing the new optimal path, and receives a higher return in the new environment.
It empirically implies that policies receiving higher returns are supposed to be more in line with the new environment, and hence contain more new knowledge.

Based on the above observation, a particular instance's quality $Qu(\pi_{\bm{z}})$ can be directly approximated by the received return of its induced policy as:
\begin{equation}
Qu(\pi_{\bm{z}}) = r(\pi_{\bm{z}}).
\end{equation}
And the second metric for assigning the instance weight is calculated as:
\begin{equation}
w(\bm{z}_i) = m\cdot\frac{e^{ Qu(\pi_{\bm{z}_i})/\rho}}{\sum_{j=1}^me^{ Qu(\pi_{\bm{z}_j})/\rho}},~~i=1,...,m.
\label{metric2}
\end{equation}
Similarly, we also increase the temperature $\rho$ by a small increment $\Delta \rho$ at each iteration, gradually approaching the form of uniform weighting as the evolution proceeds.
During parameter updating, higher importance weights are assigned to episodes that contain more new information, thus encouraging the previous optimum of parameters to be faster adjusted to a new one that fits in the new environment.
It may be helpful for the algorithm to escape from those ``deceptive" regions adjacent to the parameter space of the previous optimum.

\subsubsection{Mixing Variant}
We observe the fact that, weighting by instance novelty encourages executing policies to exhibit different behaviors from those well-performed in the original environment, while weighting by instance quality strengthens the searching for well-performing policies in the new environment.
Recalling the example in Fig.~\ref{fig:metircs}-(b), we have $N(\pi_1)\!<\!N(\pi_2)\!<\!N(\pi_3)$ and $Qu(\pi_1)\!>\!Qu(\pi_2)\!>\!Qu(\pi_3)$.
Using novelty only as the weighting metric may over-value instances with bad performance (e.g., $\pi_3$), while using quality only may lead to policies getting stuck in a deceptive trap (e.g., $\pi_1$).
Therefore, to make the most use of the two metrics, we explore a mixing variant to calculate the weight as:
\begin{equation}
w(\bm{z}_i) = m\cdot\frac{e^{ N(\pi_{\bm{z}_i})\cdot Qu(\pi_{\bm{z}_i})/\rho}}{\sum_{j=1}^me^{ N(\pi_{\bm{z}_j})\cdot Qu(\pi_{\bm{z}_j})/\rho}},~~i=1,...,m.
\label{mix}
\end{equation}
The following experimental results show that incremental evolution strategies with the mixing weighting metric generally perform the best among all compared implementations.

\subsection{Integrated Algorithm}
Based on the problem formulation and the instance weighting mechanism with designed metrics, Algorithm~\ref{ies} presents the integrated IW-IES algorithm for RL in dynamic environments.
In the incremental learning setting, an RL agent is interacting with a dynamic environment $\mathcal{D}=[M_1,M_2,...]$.
In the first time period, the policy parameters are randomly initialized in Line 3, followed by the evolutionary process from scratch using the canonical NES algorithm in Lines 4-14.
In a subsequent $t$-th ($t\ge 2$) time period (new environment), we first obtain the previous optimal policy parameters $\bm{\theta}_{t-1}^*$ and the corresponding behavior characterization $b(\pi_{\bm{\theta}_{t-1}^*})$ from the last time period (original environment) in Line 16.
Then, the policy parameters are initialized from the previous optimum in Line 17.
In Line~22, we assign a weight to each instance according to the designed metrics, aiming at facilitating the learning adaptation to the new environment.
Finally, the policy parameters iteratively evolve in Line~27 until the new optimum $\bm{\theta}_t^*$ is obtained for $M_t$.  

The use of specific metrics for the instance weighting mechanism yields three variants of implementations: 1) \textit{IW-IES-N}: using instance novelty to calculate weights in~(\ref{metric1}); 2) \textit{IW-IES-Qu}: using instance quality as the weighting metric in~(\ref{metric2}); 3) \textit{IW-IES-Mix}: using the mixing weighting metric in~(\ref{mix}).
Instance novelty is designed to measure the instance's difference from the previous optimum of the original environment, while instance quality corresponds to how well the instance performs in the new environment.
Together, an instance with high novelty or high quality is supposed to contain more knowledge on the new environment.
With this mechanism, IW-IES prefers behaviors that either differ more from the original environment or better fit in the new environment, thus encouraging the search distribution to move towards new promising regions of parameter space.

\begin{algorithm}[tb]
\caption{Instance Weighted Incremental Evolution Strategies (IW-IES)} \label{ies}
\KwIn{Dynamic environment $\mathcal{D}\!=\!\{M_1,M_2,...\}$; \newline 
current time period $t (t\ge 1)$; \newline 
learning rate $\alpha$; population size $m$; \newline 
noise standard deviation $\sigma$; \newline
increment of temperature $\Delta \rho$}
\KwOut{Optimal policy parameters $\bm{\theta}_t^*$ for $M_t$}
Initialize: $m$ CPU workers with known random seeds \\
\eIf{t equals to 1}
    {Randomly initialize $\bm{\theta}_t$ \\
    \While{not converged}{
    	\For{\text{each worker} $i=1,...,m$}{
    		Sample $\bm{\epsilon}_i\sim\mathcal{N}(\bm{0}, \bm{I})$ \\
    		Compute fitness $f_i=f(\bm{\theta}_t+\sigma\bm{\epsilon}_i)$
    	}	
    	Send all scalar fitness $f_i$ from each worker to every other worker \\
    	\For{\text{each worker} $i=1,...,m$}{
    		Reconstruct all perturbations $\bm{\epsilon}_i$ using known random seeds \\
    		Set $\bm{\theta}_t \leftarrow \bm{\theta}_t + \alpha\frac{1}{m\sigma}\sum_{i=1}^mf_i\bm{\epsilon}_i$
    	}
        }}
    {Obtain $\bm{\theta}_{t-1}^*$ and behavior characterization $b(\pi_{\bm{\theta}_{t-1}^*})$ \\
     Initialize: $\bm{\theta}_t \leftarrow \bm{\theta}_{t-1}^*$, and the temperature $\rho$ \\
    \While{not converged}{
    	\For{\text{each worker} $i=1,...,m$}{
    		Sample $\bm{\epsilon}_i\sim\mathcal{N}(\bm{0}, \bm{I})$ \\
    		Compute fitness $f_i=f(\bm{\theta}_t+\sigma\bm{\epsilon}_i)$ \\
    		Calculate instance weights $w_i\!=\!w(\bm{\theta}_t+\sigma\bm{\epsilon}_i)$ by metrics in (\ref{metric1}), (\ref{metric2}), or (\ref{mix})
    	}	
    	Send all scalar fitness $f_i$ and weights $w_i$ from each worker to every other worker \\
    	\For{\text{each worker} $i=1,...,m$}{
    		Reconstruct all perturbations $\bm{\epsilon}_i$ using known random seeds \\
    		Set $\bm{\theta}_t \leftarrow \bm{\theta}_t + \alpha\frac{1}{m\sigma}\sum_{i=1}^mw_if_i\bm{\epsilon}_i$
    	}
        $\rho\leftarrow\rho+\Delta\rho$
        }
    }
\end{algorithm}

\begin{remark}[Scalability]
As shown in~\citep{salimans2017evolution}, ES scales well with the amount of computation available, enabling a near-linear speedup in runtime as more CPUs are used.
The proposed algorithm, IW-IES, enjoys the same parallelization benefits as ES because it uses an almost identical optimization process.
In IW-IES, broadcasting both scalars of fitness $f(\bm{\theta}_t+\sigma\bm{\epsilon}_i)$ and instance weight $w(\bm{\theta}_t+\sigma\bm{\epsilon}_i)$ would incur almost zero extra network overhead, because the scalars usually take up much less memory than the large parameter vector $\bm{\theta}_t$ that must be broadcast at the beginning of each iteration.
Moreover, the addition of the behavior characterization of previous optimal policy does not hurt scalability, because it is kept fixed during the calculation of instance novelty and the coordinator needs to broadcast it only once at the beginning of each iteration.
\end{remark}

\begin{remark}[Complexity]
Here we also give a rough complexity analysis on the designed metrics.
First, for calculating instance novelty $N(\pi_{\bm{z}})$, we need to compute each instance's behavior characterization $b(\pi_{\bm{z}})$, which can be simultaneously acquired when primitively executing the associated behavior policy to compute its fitness.
For example, in the humanoid locomotion case, the observed reward and state can be used for calculating the fitness and the behavior characterization, respectively.
Besides, the behavior characterization usually consumes much less memory than the parameter vector of instance $\bm{z}$.
Second, calculating instance quality $Qu(\pi_{\bm{z}})$ would not consume any computation for extra variables, since the quality indicator, i.e., the received return $r(\pi_{\bm{z}})$, has been already acquired in the primitive step for computing the fitness.
Together, it can be inferred that the designed weighting metrics hardly increase the computational complexity of ES.
Hence, we claim that the two metrics are ``easy-to-implement".
\end{remark}

\section{Experiments}
To test IW-IES, we conduct experiments on challenging RL tasks ranging from classical navigation tasks to benchmark MuJoCo robot locomotion tasks~\citep{todorov2012mujoco}.
Using agents in these tasks, we design challenging dynamic environments that involve (multiple) changes in the underlying environment distribution, where incremental learning is critical.
Through these experiments, we aim to build problem settings that are representative types of disturbances and shifts that a real RL agent may encounter in practical applications.
The questions that we aim to study from our experiments include: 
\begin{itemize}
\item[Q1] Can IW-IES handle various dynamic environments where the reward or state transition function changes over time?
\item[Q2] Can IW-IES successfully facilitate rapid learning adaptation to these dynamic environments?
\item[Q3] How do the two weighting metrics, instance novelty and instance quality, affect the performance of IW-IES, respectively?
\end{itemize}

\subsection{Experimental Settings}
We compare IW-IES to four baseline methods: 
\begin{enumerate}
	\item \textit{Robust}~\citep{sheckells2019using}: It takes the most recent observation (i.e., $\pi_{robust}: s\mapsto a$) as input and uses domain randomization to train a robust policy that is supposed to work for all training environments, while current environmental dynamics cannot be identified from its input.
	\item \textit{SO-CMA}~\citep{yu2019policy}: It uses the environment feature $\mu$ as additional input (i.e., $\pi_{so}: [s, \mu]\mapsto a$) and trains an environment-conditioned policy  with domain randomization.
	Given particular $\mu$, the instantiated policy is called a strategy.
	In the target environment, it performs Strategy Optimization using CMA-ES, which only optimizes the environment feature input to the policy.
	\item \textit{Hist}~\citep{peng2018sim}: The adaptive policy is represented as a long short-term memory (LSTM) network that takes a history of observations as input, i.e., $\pi_{adapt}: [s_{t-h},...,s_t]\mapsto a$.
	This allows the policy to implicitly identify the environment being tested and to adaptively choose actions based on the identified environment.
	\item \textit{ES-MAML}~\citep{song2020esmaml}: It trains a meta-policy on a variety of tasks based on the NES algorithm, such that it can solve new learning tasks using only a small number of training samples. More details about MAML can be found in~\citep{finn2017model}.
\end{enumerate}

In all domains, the policy model evolved by IW-IES is instantiated as a feedforward neural network with two $128$-unit hidden layers separated by ReLU nonlinearities, similar to the benchmark network architectures used in~\citep{salimans2017evolution,conti2018improving}.
For fair comparison to our method, the network architecture of Robust and ES-MAML is set the same as that of IW-IES.
SO-CMA uses the same network architecture except that it takes the environment feature as additional input.
For Hist, we feed a history of $5$ observations to a recurrent policy network that consists of a $64$-unit embedding layer and a $64$-unit LSTM layer separated by ReLU nonlinearities.
In this way, the recurrent policy network has the same order of magnitude number of parameters as the policy model of IW-IES.
While existing methods mostly use policy gradient algorithms to train neural network policies, we implement these baseline methods by the NES algorithm as a more challenging reference point.
The number of CPU workers is set as $m=16$ for parallelizing IW-IES and all baseline methods.
The universal polices of baselines are trained over a variety of environments that are randomly sampled from  a known distribution.
For SO-CMA, the environment feature is assumed to essentially capture the MDP drawn from the distribution, which for instance can be represented by the position of goals or obstacles in a navigation task.
Further, we continue to train the universal polices after transferring to the new task whenever the environment changes, using the same number of samples IW-IES consumes in each environment.
We refer to this additional training step as ``fine-tuning". 
In contrast, IW-IES focuses on directly adapting to dynamic environments on the fly, avoiding access to a large distribution of training environments and releasing the dependency on structural assumptions of environmental dynamics.

For each report unit (a particular algorithm running on a particular task), we define two performance metrics.
One is the received return for executing the policy induced by the unperturbed instance in each evolution generation, defined as $r(\pi_{\bm{\theta}})$.
The other is the average return received over all generations, defined as $\frac{1}{I}\sum_{i=1}^Ir_i(\pi_{\bm{\theta}})$, where $I$ is the number of total evolving generations.
The former will be plotted in figures and the latter will be presented in tables.
Due to the randomness of training neural networks, we run three trials with different seeds and adopt the mean as the performance for each report unit.
We utilize a statistical analysis method to address the issue of ``stochastic" dynamic environments.
The learning agent first learns an optimal policy given a randomly chosen environment.
Then, the environment randomly changes to a new one, and we record the performance of all tested methods when adapting to the new environment.
We repeat the process for $10$ times and report the mean and standard error to demonstrate the performance for learning in stochastic dynamic environments.
The code is available online.
\footnote{\href{https://github.com/HeyuanMingong/iwies}{https://github.com/HeyuanMingong/iwies}}


\subsection{Navigation Tasks}
We first test our IW-IES algorithms on a set of navigation tasks where a point agent must move to a goal position in 2D within a unit square. 
The state is the current observation of the 2D position, and the action corresponds to the 2D velocity commands that are clipped to be in the range of $[-0.1,0.1]$. 
The reward is the negative squared distance to the goal minus a small control cost that is proportional to the action's scale.
Each learning episode (generated by the instance $\bm{z}$) always starts from a given point and terminates when the agent is within $0.01$ of the goal or at the horizon of $H=100$.
The hyperparameters are set as: population size $m=16$ and noise standard deviation $\sigma=0.05$.
The learning rate is set as the same for all tested methods in each task.

As described in Section~III-C, the first weighting metric requires a domain-specific behavior characterization for calculating the novelty of each instance $\bm{z}$.
For the navigation problems, the behavior characterization is the trace of the agent's $(x,y)$ locations through all time steps:
\begin{equation}
b(\pi_{\bm{z}}) = \{(x_{\bm{z}}^1,y_{\bm{z}}^1),...,(x_{\bm{z}}^H,y_{\bm{z}}^H)\}.
\end{equation}
Computing instance novelty also requires a distance function between behavior characterizations of the instance $\bm{z}$ and the previous optimum $\bm{\theta}_{t-1}^*$. 
Following~\citep{conti2018improving,lehman2011abandoning}, we use the average Euclidean distance of these 2D coordinates as the distance function:
\begin{equation}
\dist(b(\pi_{\bm{z}}),b(\pi_{\bm{\theta}_{t-1}^*})) \!\!=\!\! \frac{1}{H}\!\sum_{i=1}^H\!\sqrt{(x_{\bm{z}}^i \!-\! x_{t-1}^i)^2 \!+\! (y_{\bm{z}}^i \!-\! y_{t-1}^i)^2},
\end{equation}
where $\{(x_{t-1}^i,y_{t-1}^i)\}_{i=1}^H$ are the agent's locations when executing the previous optimal policy $\pi_{\bm{\theta}_{t-1}^*}$.

\subsubsection{Q1}
\begin{figure}[tb]
\centering
\subfigure[]{\includegraphics[width=0.95\columnwidth]{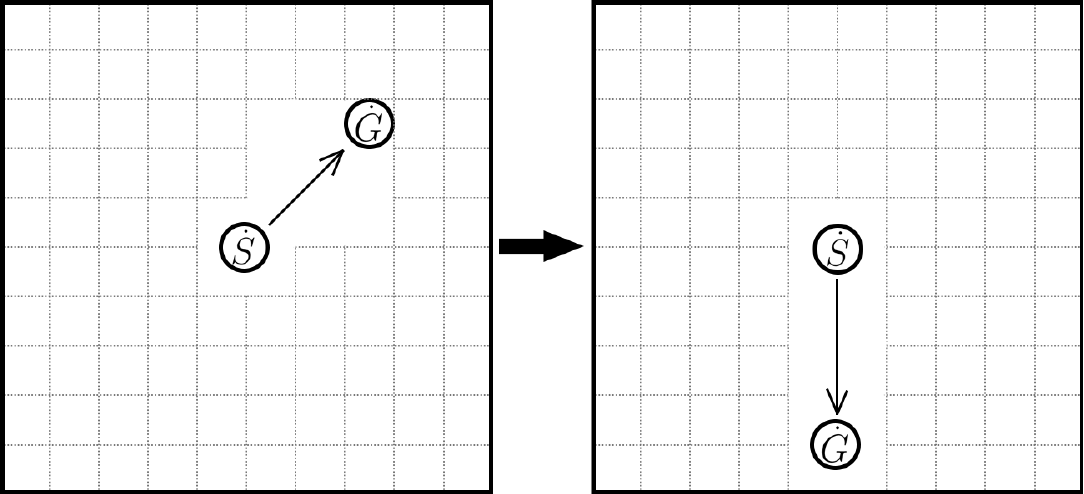}}
\subfigure[]{\includegraphics[width=0.95\columnwidth]{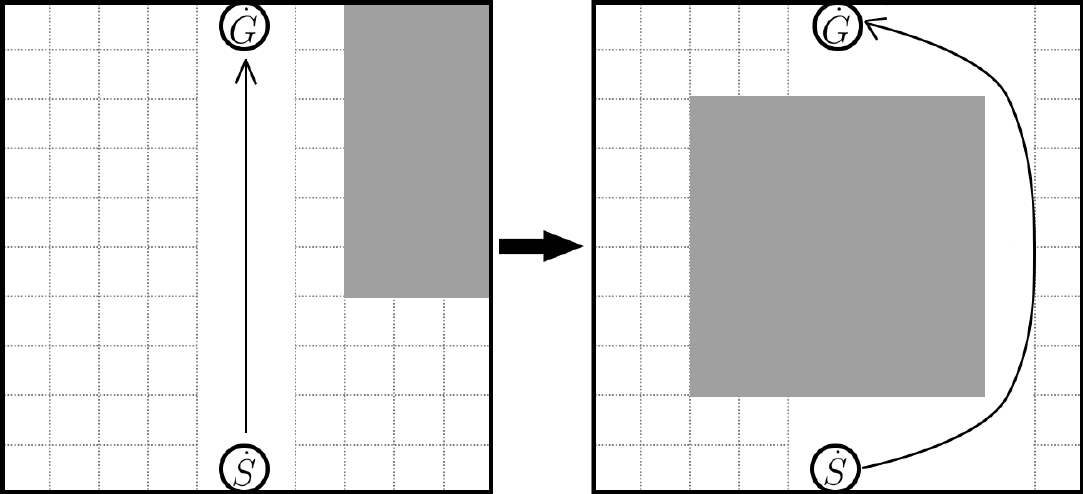}}
\caption{Two illustrative cases of dynamic environments in the 2D navigation tasks. $\dot{S}$ is the start point and $\dot{G}$ is the goal point. 
(a) Case I: the goal changes. (b) Case II: the landscape changes, and the gray is a square obstacle.}
\label{fig:2example}
\end{figure}

We start with two illustrative cases of simulated dynamic environments as shown in Fig.~\ref{fig:2example}: 
\begin{itemize}
\item[(a)] \textbf{Case I:} The dynamic environment is created by randomly changing the goal position, while keeping the start point fixed at $(0,0)$. 
The environment changes in terms of the reward function, and the goal position can be used as the environment feature for SO-CMA.  
\item[(b)] \textbf{Case II:} The start and goal points are kept fixed at $(0,-0.5)$ and $(0,0.5)$, respectively. 
The dynamic environment is created by moving a $0.6\times 0.6$ square obstacle at random.
When hitting on the obstacle, the agent will bounce to its previous position.
The environment changes in terms of the state transition function, and the environment feature can be represented by the centering position of the square obstacle.
\end{itemize}

\subsubsection{Q2}
We present the primary experimental results of baselines and IW-IES implemented on the two cases of dynamic environments.
The average return (of the executing policy induced by the unperturbed instance) per generation across $10$ independent runs is plotted in Fig.~\ref{fig:rews}.
Here and in similar figures below, the mean of average return per generation across 10 runs is plotted as the bold line with 95\% bootstrapped confidence intervals of the mean (shaded).
Further, Table~\ref{tab:rews} reports numerical results in terms of average received return over $200$ training generations for Case I and over $1000$ generations for Case II.
Here and in similar tables below, the mean across $10$ runs is presented, and the confidence intervals are corresponding standard errors. The best performance is marked in boldface.

\begin{figure*}[tb]
	\centering 
	\subfigure[]{\includegraphics[width=0.9\columnwidth]{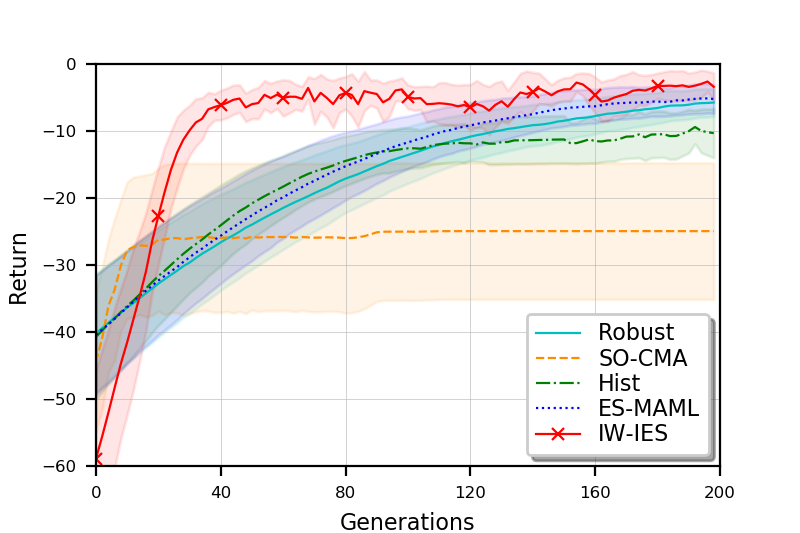}}
	\subfigure[]{\includegraphics[width=0.9\columnwidth]{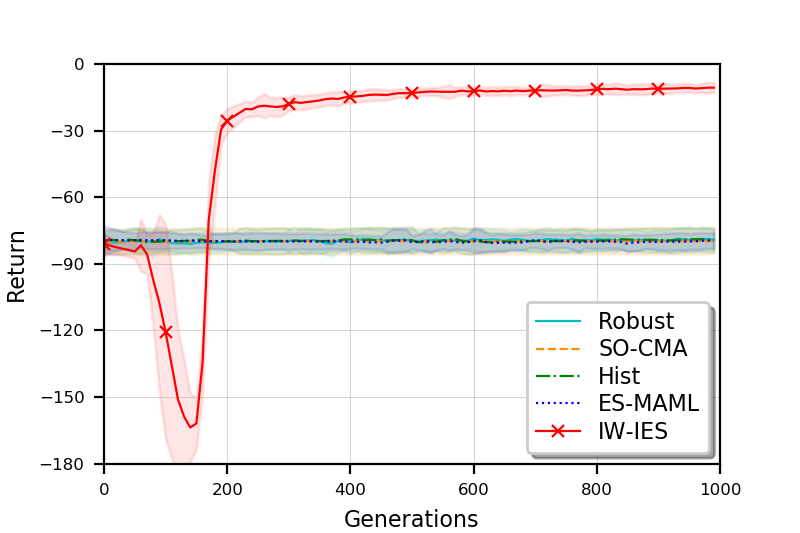}}
	\caption{The received return per generation of baselines and IW-IES implemented on two navigation tasks. (a) Case I. (b) Case II.}
	\label{fig:rews}
\end{figure*}

In Case I, since all baselines pretrain the policy model over a large distribution of randomized source environments, they receive higher \textit{jumpstart} return than IW-IES when starting interacting with the new environment.
SO-CMA obtains good jumpstart performance in the beginning.
However, in the latter learning process, it receives non-increasing return that is smaller than other baseline methods.
SO-CMA can adapt to dynamic environments with fewer data by optimizing only the environment feature input to the policy~\citep{yu2019policy}, whereas its final performance may be inferior to methods that adjust the neural network weights in the fine-tuning phase.
By comparison, in spite of obtaining smaller return initially, IW-IES exhibits significantly faster learning adaptation to dynamic environments than all baselines.
For instance, IW-IES receives near-optimal return with only $40$ generations, whereas Robust, Hist and ES-MAML need to take more than $200$ generations for achieving comparable performance.
Additionally, the statistical results show that IW-IES obtains smaller confidence intervals and standard errors than all baselines, indicating that IW-IES can provide more stable learning adaptation to new environments.

\begin{table}[tb]
	\centering
	\renewcommand\arraystretch{1.2}
	\setlength{\tabcolsep}{5.0mm}
	\caption{Average received return over all training generations of baselines and IW-IES implemented on two navigation tasks.}
	\label{tab:rews}
	\begin{tabular}{c|c|c}
		\cmidrule[\heavyrulewidth]{1-3}
		Methods & Case I & Case II \\
		\hline 
		Robust         & $-16.99\pm 2.08$             & $-79.59\pm 1.10$  \\
		SO-CMA      & $-25.99\pm 4.57$           & $-79.79\pm 1.40$  \\
		Hist      & $-17.22\pm 1.85$             & $-79.55\pm 1.19$  \\
		ES-MAML    & $-15.69\pm 2.03$             & $-79.90\pm 0.86$  \\
		IW-IES         & $\bm{-8.65\pm 1.04}$    & $\bm{-31.43\pm 0.77}$ \\
		\cmidrule[\heavyrulewidth]{1-3}
	\end{tabular}
\end{table}

Next, the results in Case II reveal some differences.
In Case I, IW-IES achieves faster learning adaptation than baselines, while the performance gap is relatively moderate.
The reason is that, in this simple case of navigation, all baselines can evolve near-optimal policies that will find a path to the new goal as shown in Fig.~\ref{fig:2example}-(a).
In contrast, IW-IES obtains much higher final return in Case II.
Since navigating to the goal while bypassing the wall is usually difficult, policies evolved by baselines tend to head directly towards the goal.
Thus, it easily terminates in front of the huge obstacle and gets stuck in bad local optima.
Instead, the proposed instance weighting mechanism ``reinforces" the policy evolved by IW-IES to explore behaviors that are different from previous optima (instance novelty) and to exploit behaviors that are in line with the new environment (instance quality).
Therefore, the policy evolved by IW-IES is more capable of bypassing the deceptive wall first and moving to the goal finally. 
The primary results verify that the instance weighting mechanism effectively encourages the search distribution to move towards promising regions of parameter space that fit in the new environment.
IW-IES can obtain superior performance compared to all baselines, demonstrating the effectiveness of our method for addressing incremental learning problems in dynamic environments.

\begin{figure}[tb]
	\centering 
	\includegraphics[width=0.9\columnwidth]{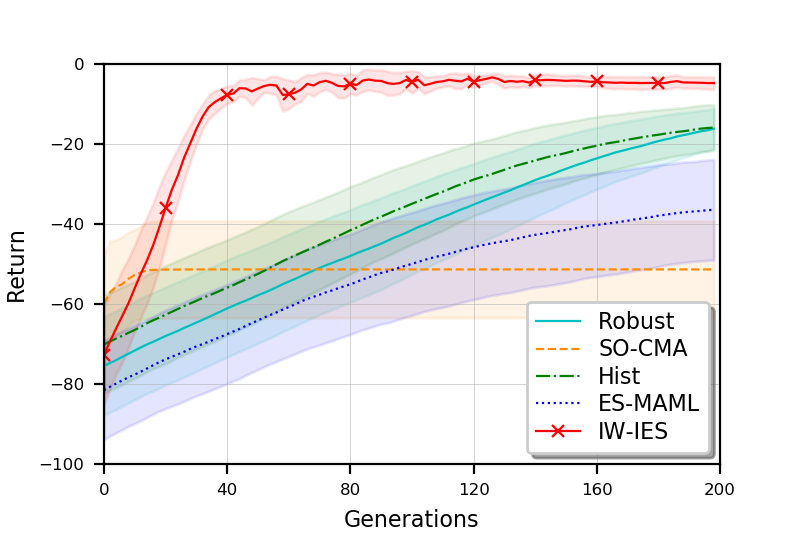}
	\caption{Received return per generation of baselines and IW-IES in Case I navigation task with a large discrepancy between environment distributions.}
	\label{fig:rews_dis}
\end{figure}

In the above experiments, the original and new environments are sampled from the same distribution.
In addition, we employ the Case I navigation task to serve as an illustrative example to test the performance of baselines and IW-IES when the discrepancy between distributions of the original and new environments is large.
The universal policies of baselines are trained over a limited range of environments where goal positions are in the first quadrant. 
Then, they are transferred to and fine-tuned in new environments where goal positions are in the third quadrant.
For IW-IES, the original and new environments are sampled from the same distributions as those of baselines, respectively.
Fig.~\ref{fig:rews_dis} presents the received return per generation of baselines and IW-IES, and Table~\ref{tab:rews_dis} shows corresponding numerical results.
It is observed that the advantage of IW-IES over baselines is more prominent than the case with no discrepancy of environment distributions as shown in Fig.~\ref{fig:rews}-(a). 
Consistent with the analysis in Section II-B, existing methods rely on transferring the knowledge trained over a large distribution of source tasks to new environments that are in line with the source distribution.
The universal policies trained by baselines may not generalize well when the discrepancy between distributions of the original and new environments is too large.
In contrast,  IW-IES can stably facilitate learning adaptation to dynamic environments regardless of the distribution discrepancy.

\begin{table}[tb]
	\centering
	\renewcommand\arraystretch{1.2}
	\setlength{\tabcolsep}{5.0mm}
	\caption{Average received return over all generations of baselines and IW-IES implemented on Case I navigation tasks.}
	\label{tab:rews_dis}
	\begin{tabular}{c|c|c}
		\cmidrule[\heavyrulewidth]{1-3}
		Methods & Without discrepancy & With discrepancy \\
		\hline 
		Robust         & $-16.99\pm 2.08$             & $-42.38\pm 4.72$  \\
		SO-CMA      & $-25.99\pm 4.57$           & $-51.67\pm 5.34$  \\
		Hist      			& $-17.22\pm 1.85$             & $-38.05\pm 4.26$  \\
		ES-MAML    & $-15.69\pm 2.03$             & $-53.49\pm 5.58$  \\
		IW-IES         & $\bm{-8.65\pm 1.04}$    & $\bm{-11.57\pm 1.03}$ \\
		\cmidrule[\heavyrulewidth]{1-3}
	\end{tabular}
\end{table}

\subsubsection{Q3}
\begin{figure*}[tb]
	\centering 
	\subfigure[]{\includegraphics[width=0.9\columnwidth]{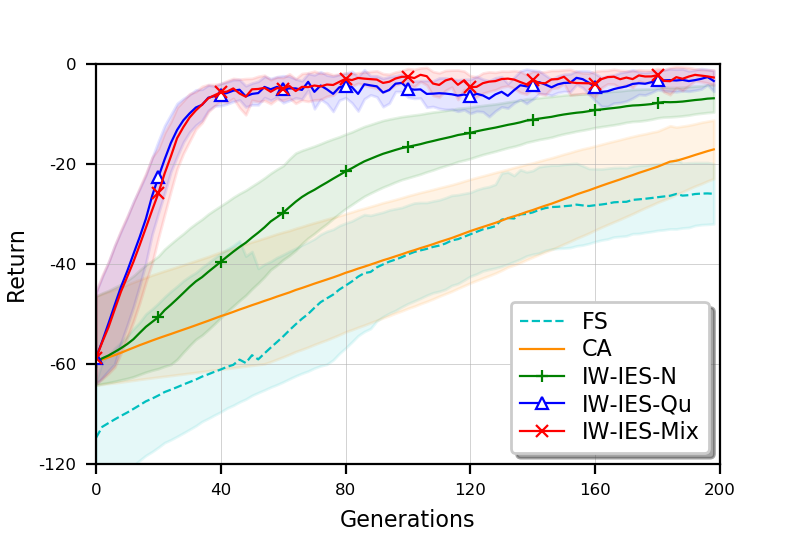}}
	\subfigure[]{\includegraphics[width=0.9\columnwidth]{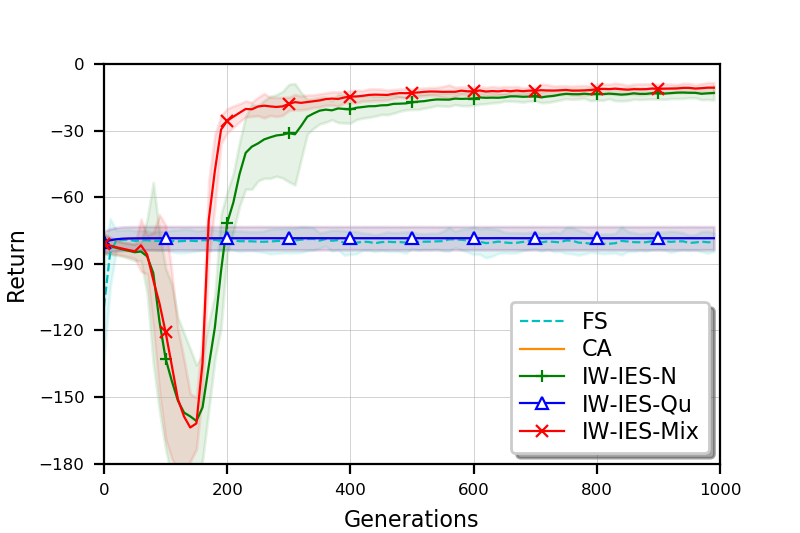}}
	\caption{The received return per generation of the four variants of IW-IES implemented on two navigation tasks. (a) Case I. (b) Case II.}
	\label{fig:iwies}
\end{figure*}

To identify the respective effects of the two weighting metrics, we adopt a control variate approach to separate them apart for observation.
In each task, we first initialize policy parameters from the original environment, and then implement four variants of IW-IES according to the employed weighting metric: 
\begin{itemize}[leftmargin=1em]
\item \textbf{CA:} No instance weighting mechanism is applied, i.e., Continuously Adapting a single policy model whenever the environment changes.
This is representative of commonly used dynamic evaluation methods~\citep{nagabandi2019learning}.
\item \textbf{IW-IES-N:} Use instance novelty in~(\ref{metric1}) as the metric.
\item \textbf{IW-IES-Qu:} Use instance quality in~(\ref{metric2}) as the metric.
\item \textbf{IW-IES-Mix:} Use the mixing weighting metric in~(\ref{mix}).
\end{itemize}
In addition, we also  investigate the performance of the policy trained with randomly initialized parameters whenever the environment changes, i.e., learning From Scratch (\textbf{FS}).
For the four variants and FS, Fig.~\ref{fig:iwies} presents their learning performance per evolution generation, and Table~\ref{tab:iwies} reports the average received return over all training generations. 

\begin{table}[tb]
	\centering
	\renewcommand\arraystretch{1.2}
	\setlength{\tabcolsep}{5.0mm}
	\caption{Average received return over all training generations of the four variants of IW-IES and FS on two navigation tasks.}
	\label{tab:iwies}
	\begin{tabular}{c|c|c}
		\cmidrule[\heavyrulewidth]{1-3}
		Methods & Case I & Case II  \\
		\hline 
		FS					  & $-49.12\pm 5.42$          & $-79.53\pm 1.06$  \\
		CA 					  & $-37.94\pm 4.70$         & $-78.48\pm 1.26$  \\
		IW-IES-N 		& $-23.53\pm 2.97$           & $-38.59\pm 1.57$ \\
		IW-IES-Qu 	  & $-9.34\pm 0.93$            & $-78.50\pm 1.26$  \\
		IW-IES-Mix    & $\bm{-8.65\pm 1.04}$  & $\bm{-31.43\pm 0.77}$  \\
		\cmidrule[\heavyrulewidth]{1-3}
	\end{tabular}
\end{table}

We observe that CA usually has better performance than FS regarding the adaptation speed and the average received return.
Especially, CA can achieve jumpstart performance at the beginning of the new learning process compared to FS.
It is consistent with the analysis in Section III-B that, initializing policy parameters from the original environment empirically benefits the evolutionary process since the previous optimum has learned some of the feature representations of the state-action space.
This is also a basic impetus for the formalized incremental learning procedure in the paper.

In Case I, using instance novelty or instance quality alone as the weighting metric can achieve faster learning adaptation to the dynamic environment.
The performance gap in terms of average return is more pronounced for smaller amount of computation, which is supposed to benefit from the instance weighting mechanism that allows for distinct acceleration of incremental learning adaptation.
The weighting metric using instance quality (IW-IES-Qu) better improves the learning performance than the one using instance novelty (IW-IES-N), and combining the two weighting metrics together (IW-IES-Mix) achieves slightly better performance than IW-IES-Qu.

In Case II, both CA and IW-IES-Qu fail to find the new goal when the huge obstacle blocks the previous optimal path.
In contrast, introducing the weighting mechanism with instance novelty will degrade the learning performance in initial generations.
To bypass the obstacle in this ``deceptive" case, both IW-IES-N and IW-IES-Mix need to encourage the learning agent to exhibit novel behaviors to a large extent, thus resulting in the temporarily pessimistic performance in the early stage.
As the evolution proceeds, the effect of instance weighting becomes weaker, and they can gradually find the optimal path to the new goal instead of getting stuck in front of the wall.
In this case, IW-IES-N better improves the learning performance than IW-IES-Qu, and IW-IES-Mix obtains the best learning adaptation to the dynamic environment where the obstacle changes over time.

In summary, the two weighting metrics of instance novelty and instance quality can provide distinguished advantages for different kinds of dynamic environments.
Under most circumstances, combining the two weighting metrics together leads to the best learning adaptation.

\begin{table}[tb]
\centering
\renewcommand\arraystretch{1.2}
\setlength{\tabcolsep}{3.0mm}
\caption{The runtime (seconds) of the four variants of IW-IES with varying numbers of CPU workers on Case I navigation task.}
\label{tab:time}
\begin{tabular}{c|c|c|c|c|c}
	\cmidrule[\heavyrulewidth]{1-6}
	\# CPU workers 	& 1 & 2 & 3 & 4 & 5   \\
	\hline 
	CA     		& 141.84 & 81.82 & 59.04 & 49.88 & 41.21 \\
	IW-IES-N    & 141.23 & 83.72 & 58.72 & 49.92 & 40.77 \\
	IW-IES-Qu   & 141.51 & 82.23 & 58.65 & 50.14 & 40.94 \\
	IW-IES-Mix  & 141.11 & 82.87 & 57.81 & 48.87 & 41.09  \\
	\cmidrule[\heavyrulewidth]{1-6}
\end{tabular}
\end{table}

Further, we employ the Case I navigation task to serve as an illustrative example to verify the scalability of IW-IES.
Table~\ref{tab:time} shows the runtime of the four variants of IW-IES with varying numbers of parallelized CPU workers.
It is observed that IW-IES retains scalability and enables a near-linear speedup in runtime as more CPU workers are used.
Obviously, introducing the instance novelty or instance quality as the weighting metric hardly increases the runtime of our method.  
The result empirically verifies the claim in Remark 4 that the designed weighting metrics hardly increase the computational complexity of ES.

\subsubsection{Complex Stochastic Dynamic Environments}
\begin{figure}[tb]
	\centering
	\includegraphics[width=0.95\columnwidth]{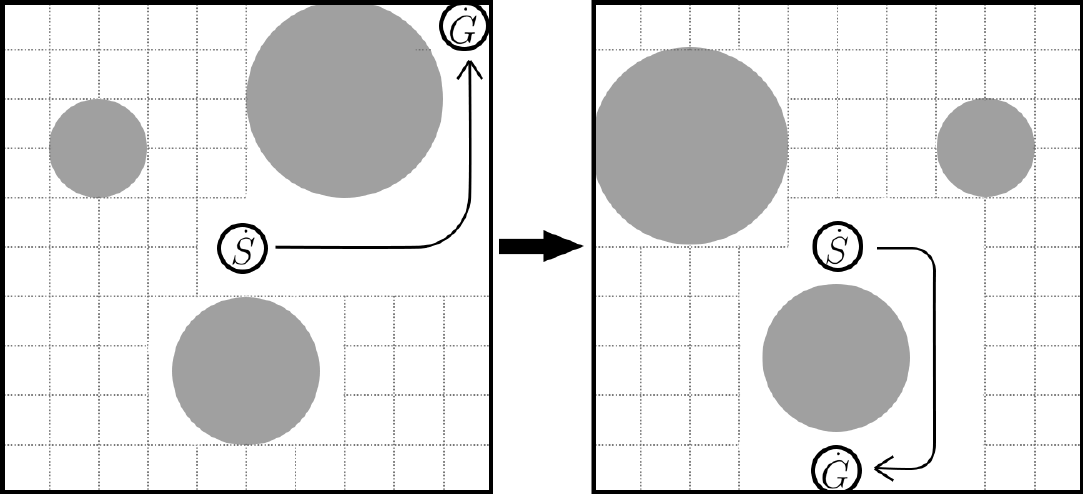}
	\caption{The 2D navigation task in a complex stochastic dynamic environment, where both the goal and circular puddles may change over time.}
	\label{fig:complex}
\end{figure}

\begin{figure}[tb]
	\centering 
	\subfigure[]{\includegraphics[width=0.49\columnwidth]{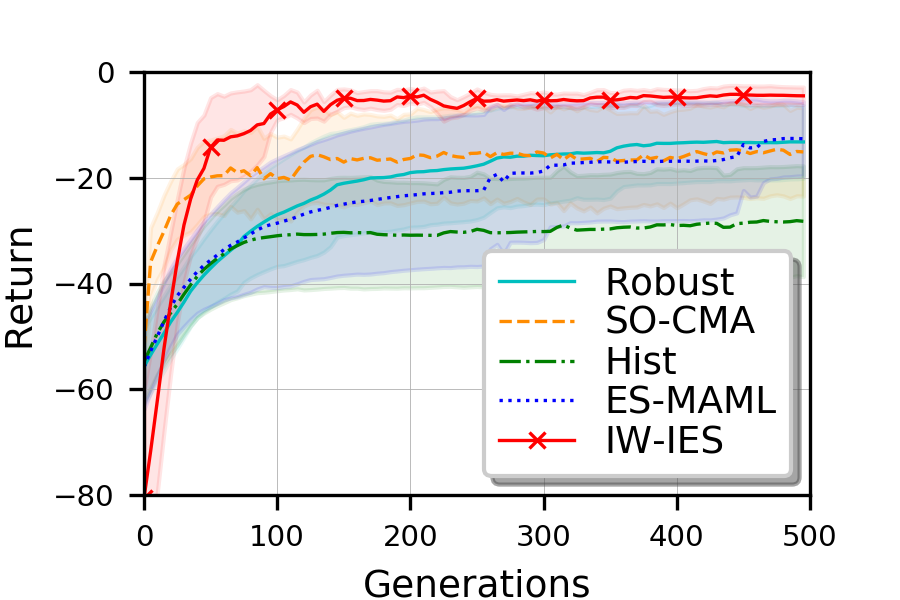}}
	\subfigure[]{\includegraphics[width=0.49\columnwidth]{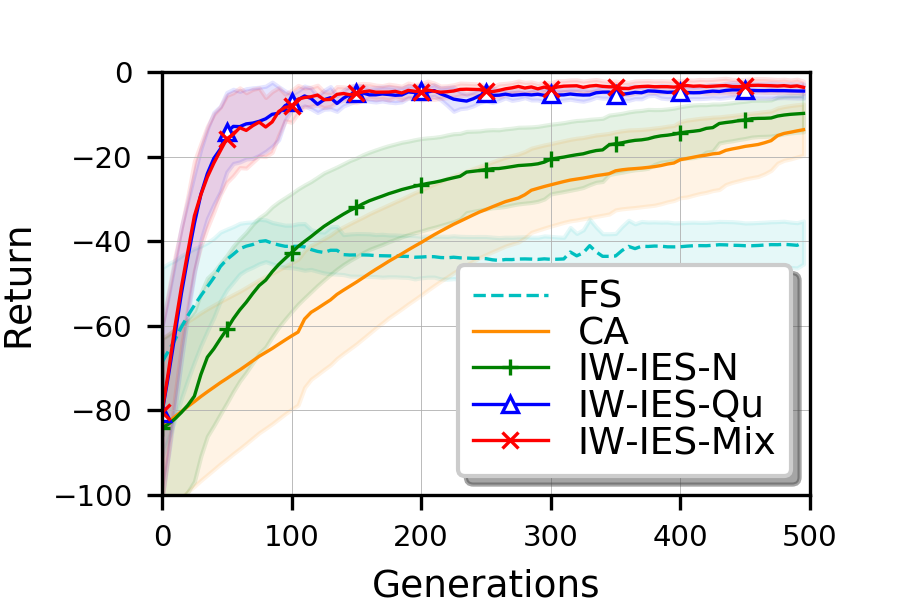}}
	\caption{The received return per generation of all tested methods implemented on the complex stochastic dynamic environment. (a) Baselines and IW-IES. (b) FS and four variants of IW-IES.}
	\label{fig:com_rews}
\end{figure}

It is verified from the above results that IW-IES can successfully facilitate the learning adaptation to dynamic environments where the reward or state transition function may change over time.
Here, we test IW-IES on a more complex case of dynamic environments, which is also a modified version of the benchmark puddle world environment presented in~\citep{sutton1996generalization,tirinzoni2018importance}.
As shown in Fig.~\ref{fig:complex}, the agent should drive to the goal while avoiding three circular puddles with different sizes.
When hitting on the puddles, the agent will bounce to its previous position.
The dynamic environment is created by moving both the goal and puddles within the unit square randomly.
Hence, the environment changes in terms of both the reward and state transition functions.
Fig.~\ref{fig:com_rews} shows the received return per generation of the baselines, FS, and the four IW-IES variants, and Table~\ref{tab:case3} presents corresponding numerical results in terms of average return over $500$ evolving generations of all tested methods.
We can observe that IW-IES is still capable of achieving significantly faster learning adaptation to this complex stochastic dynamic environment in a statistical sense.

\begin{table}[tb]
	\centering
	\renewcommand\arraystretch{1.2}
	\setlength{\tabcolsep}{2.5mm}
	\caption{Average return over $500$ evolving generations of all tested methods on the complex stochastic dynamic environment.}
	\label{tab:case3}
	\begin{tabular}{c|c|c|c}
		\cmidrule[\heavyrulewidth]{1-4}
		Methods & Average return  & Methods & Average return  \\
		\hline 
		Robust & $-21.42\pm 3.63$  & FS & $-42.91\pm 2.23$  \\
		SO-CMA & $-17.71\pm 3.54$  & CA & $-39.67\pm 4.72$ \\
		Hist & $-31.65\pm 4.32$  & IW-IES-N & $-29.97\pm 3.77$\\
		ES-MAML & $-23.46\pm 4.74$ & IW-IES-Qu & $-9.70\pm 0.85$ \\
		IW-IES & $\bm{-8.85\pm 1.06}$ & IW-IES-Mix & $\bm{-8.85\pm 1.06}$ \\
		\cmidrule[\heavyrulewidth]{1-4}
	\end{tabular}
\end{table}

From the above comprehensive experimental results on 2D navigation tasks, it can be well demonstrated that: 
\begin{itemize}
\item[A1] IW-IES is able to handle various dynamic environments that change in terms of the reward function (Fig.~\ref{fig:2example}-(a)), or the state transition function (Fig.~\ref{fig:2example}-(b)), or both (Fig.~\ref{fig:complex}).
\item[A2] IW-IES successfully enables faster and more stable learning adaptation to these dynamic environments.
\item[A3] The two weighting metrics of instance novelty and instance quality can offer distinct superiority for incremental learning in different kinds of dynamic environments, and combining them together usually achieves the best learning adaptation.
\end{itemize}

\subsection{Locomotion Tasks}
\begin{figure}[tb]
\centering 
\subfigure[]{\includegraphics[width=0.32\columnwidth]{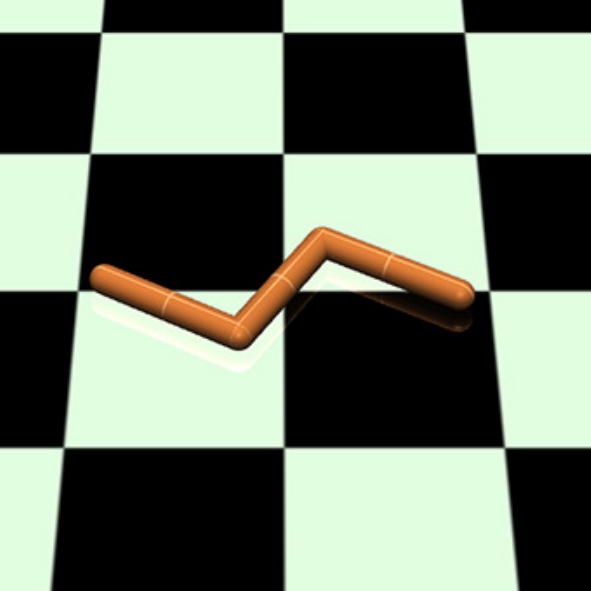}}
\subfigure[]{\includegraphics[width=0.32\columnwidth]{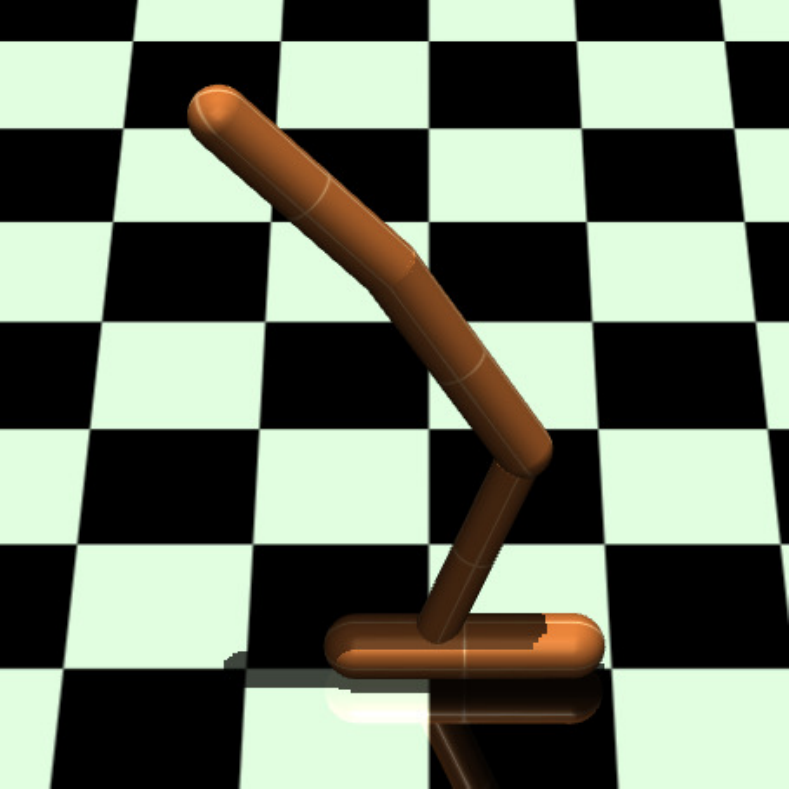}}
\subfigure[]{\includegraphics[width=0.32\columnwidth]{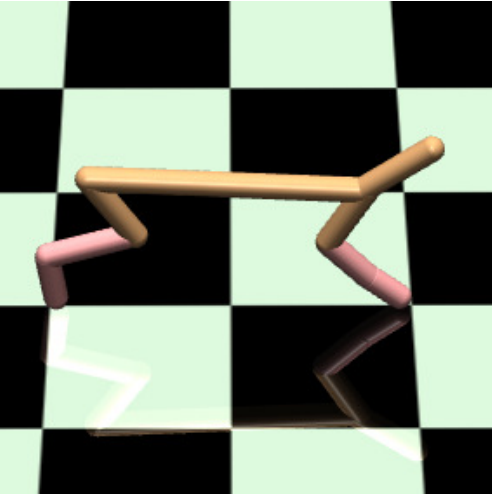}}
\caption{Challenging MuJoCo locomotion tasks including: (a) Swimmer, $|S|\!=\!8,|A|\!=\!2$; (b) Hopper, $|S|\!=\!11,|A|\!=\!3$; (c) HalfCheetah, $|S|\!=\!20,|A|\!=\!6$.}
\label{fig:envs}
\end{figure}

The above results illustrate that IW-IES is simply well suited to the 2D navigation domains, and it significantly facilitates the learning adaption to various dynamic environments.
A natural question is whether IW-IES can be successfully applied to more difficult domains.
It is necessary to test IW-IES on a well-known problem of considerable difficulty.
Thus, we also investigate three high-dimensional locomotion tasks with the MuJoCo simulator~\citep{todorov2012mujoco}, aiming at testing whether IW-IES can achieve locomotion at the scale of DNNs on much more sophisticated dynamic environments.
As shown in Fig.~\ref{fig:envs}, the continuous control tasks require a swimmer/one-legged hopper/planar cheetah robot to move at a particular velocity in the positive $x$ direction.
These three scenarios are representative locomotion tasks with growing dimensions of state and action spaces.
The reward is an alive bonus plus a regular part that is negatively correlated to the absolute value between the agent's velocity and a goal.
The goal velocity is randomly chosen between: $[0, 0.5]$ for Swimmer, $[0,1]$ for Hopper, and $[0,2]$ for HalfCheetah.
We also simulate a stochastic dynamic environment by changing the goal velocity at random across $10$ independent runs.

\begin{figure*}[tb]
	\centering 
	\subfigure[]{\includegraphics[width=0.32\textwidth]{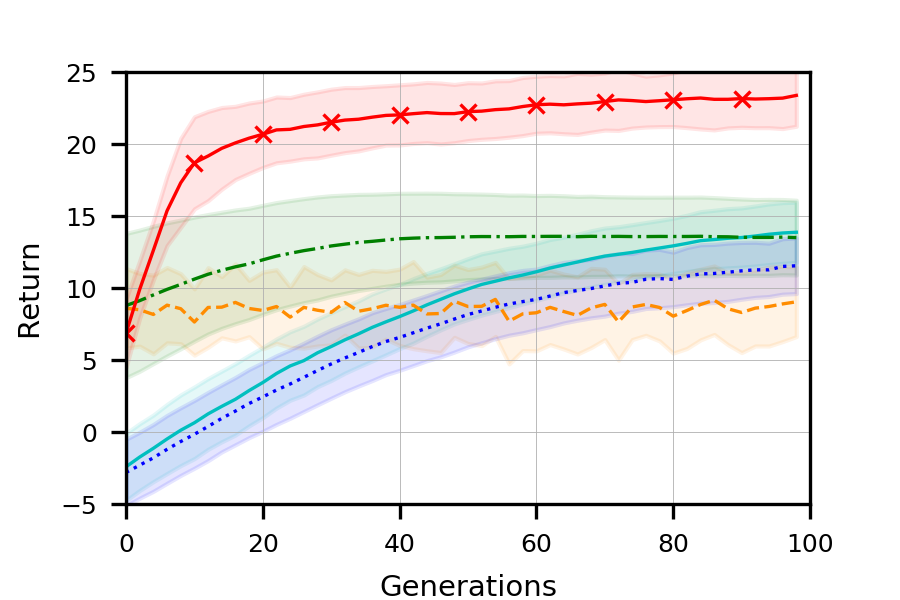}}
	\subfigure[]{\includegraphics[width=0.32\textwidth]{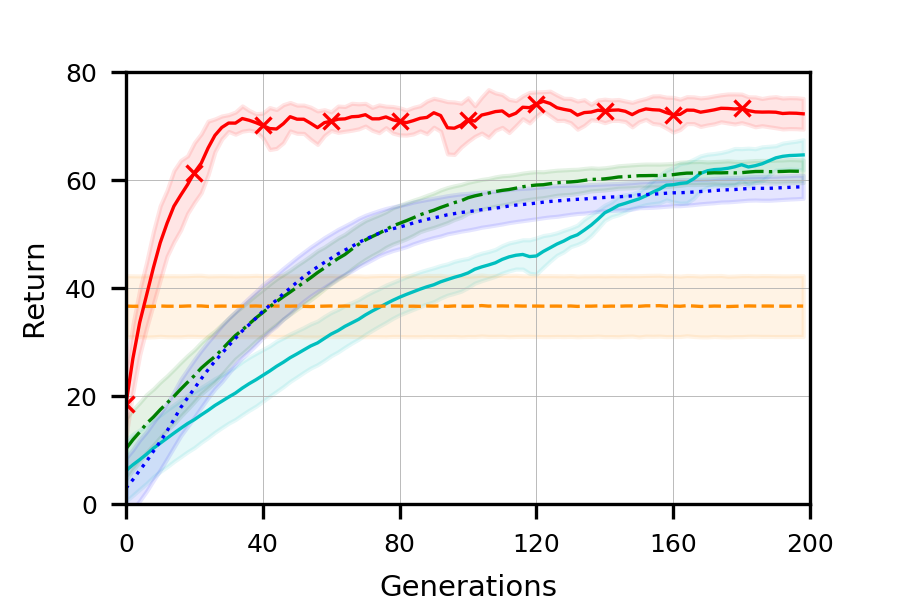}}
	\subfigure[]{\includegraphics[width=0.32\textwidth]{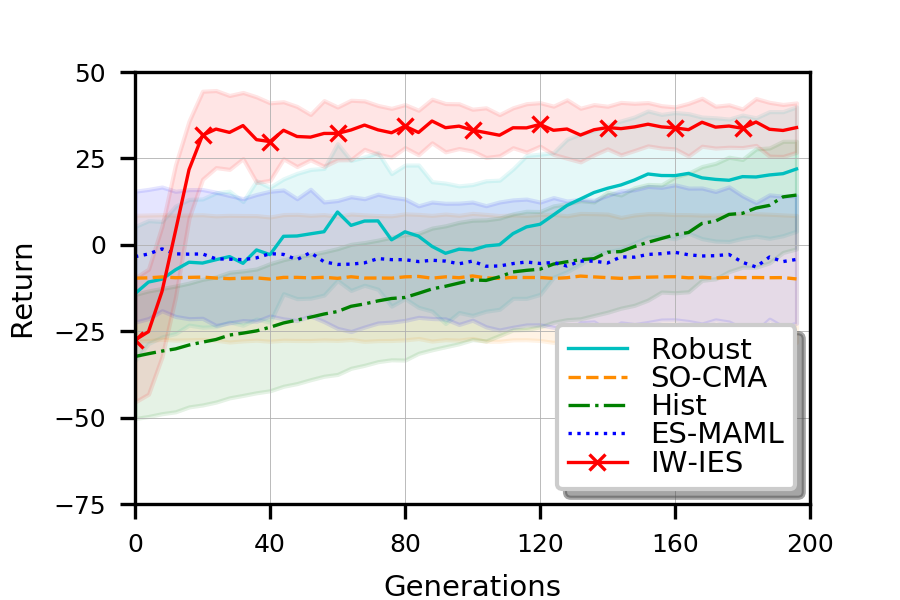}}
	\caption{The received return per generation of baselines and IW-IES on the challenging locomotion tasks. (a) Swimmer. (b) Hopper. (c) HalfCheetah.}
	\label{fig:loco_baselines}
\end{figure*}

\begin{figure*}[tb]
	\centering 
	\subfigure[]{\includegraphics[width=0.32\textwidth]{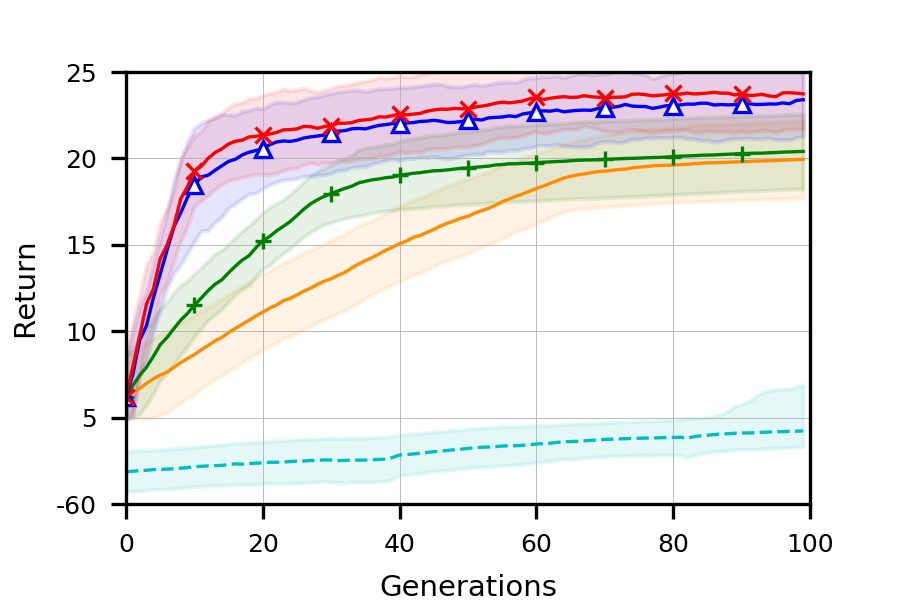}}
	\subfigure[]{\includegraphics[width=0.32\textwidth]{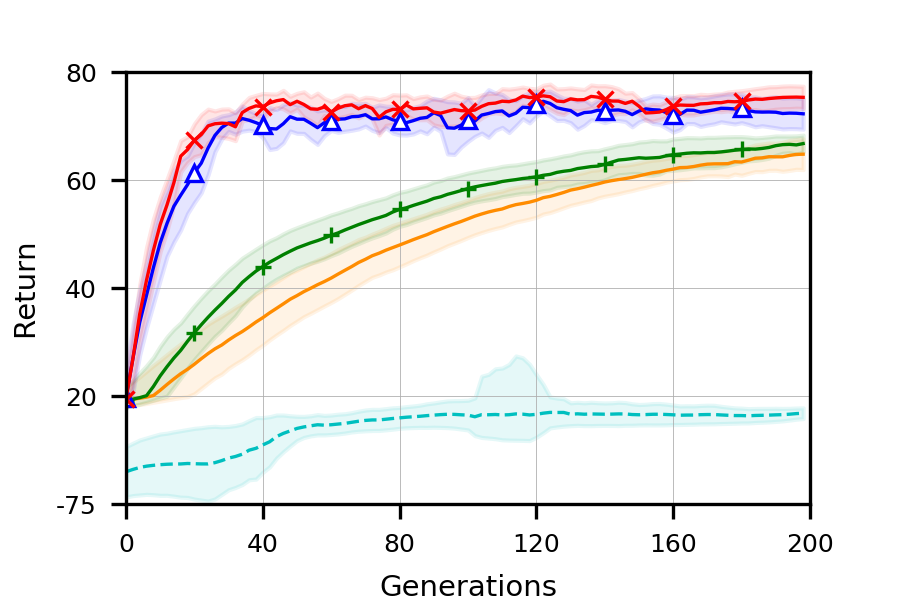}}
	\subfigure[]{\includegraphics[width=0.32\textwidth]{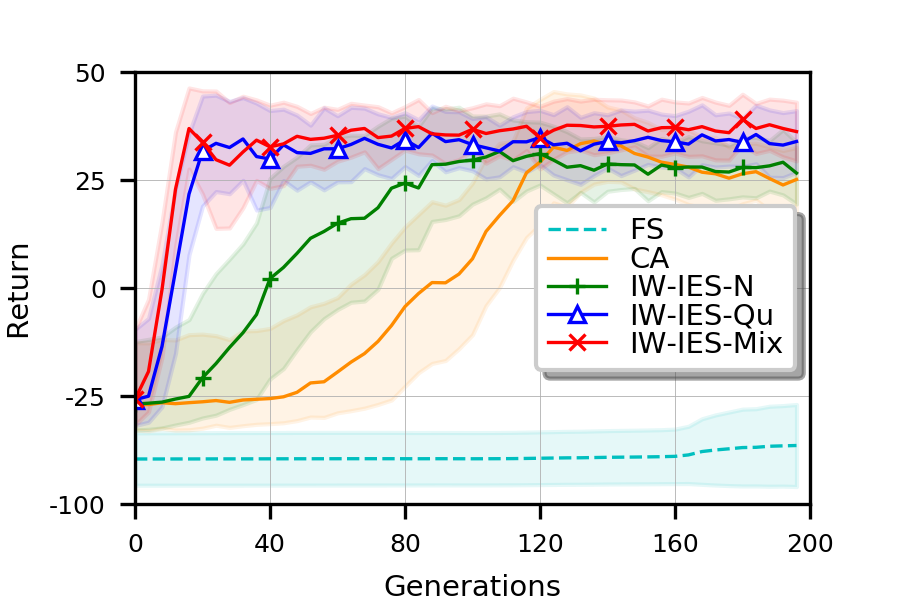}}
	\caption{The received return per generation of FS and the four variants of IW-IES on locomotion tasks. (a) Swimmer. (b) Hopper. (c) HalfCheetah.}
	\label{fig:loco_iwies}
\end{figure*}

In the locomotion domains, the behavior characterization for calculating instance novelty should be able to distinguish the robot's gaits in the $x$ direction, in accordance with the internal learning tasks.
According to this principle, a feasible behavior characterization is the offset of the robot's coordinate in the $x$ direction through all time steps:
\begin{equation}
b(\pi_{\bm{z}})=\{x_{\bm{z}}^1-x_{\bm{z}}^0,...,x_{\bm{z}}^H-x_{\bm{z}}^0\},
\end{equation}
where $x^0$ and $x^i, (i\!=\!1,...,H)$ correspond to the offsets of center of mass along the $x$ direction at the initial and the $i$-th time steps.
Consistently with the 2D navigation domains, we use the average Euclidean distance of the 1D coordinates as the distance function between behavior characterizations of instance $\bm{z}$ and the previous optimum $\bm{\theta}_{t-1}^*$:
\begin{equation}
\dist(b(\pi_{\bm{z}}),b(\pi_{\bm{\theta}_{t-1}^*})) \!=\! \frac{1}{H}\!\sum_{i=1}^H|(x_{\bm{z}}^i \!-\! x_{\bm{z}}^0) - (x_{t-1}^i \!-\! x_{t-1}^0)|,
\end{equation}
where $\{x_{t-1}^i\!-\!x_{t-1}^0\}_{i=1}^H$ is the behavior characterization associated with the previous optimal policy $\pi_{\bm{\theta}_{t-1}^*}$.

With the above settings, we present the results of all tested methods implemented on the challenging locomotion domains.
Fig.~\ref{fig:loco_baselines} shows the received return per generation of baselines and IW-IES, and Fig.~\ref{fig:loco_iwies} presents the received return per generation of FS and the four variants of IW-IES.
Corresponding numerical results in terms of average received return over all training generations are reported in Table~\ref{tab:rews_loco}.
Primarily, it is observed that IW-IES achieves more stable and faster learning adaptation in these locomotion tasks than the four baseline methods, which demonstrates the effectiveness of our method for addressing incremental learning problems in Mujoco locomotion domains.

\begin{table}[tb]
	\centering
	\renewcommand\arraystretch{1.2}
	\setlength{\tabcolsep}{2.2mm}
	\caption{Average received return over all generations of baselines, FS, and four variants of IW-IES on Mujoco locomotion tasks.}
	\label{tab:rews_loco}
	\begin{tabular}{c|c|c|c}
		\cmidrule[\heavyrulewidth]{1-4}
		Methods & Swimmer & Hopper & HalfCheetah \\
		\hline 
		Robust         & $8.22\pm 0.96$      & $40.74\pm 1.15$  & $6.56\pm 7.98$ \\
		SO-CMA      & $8.55\pm 1.00$      & $36.64\pm 2.49$ & $-9.49\pm 8.02$   \\
		Hist               & $12.75\pm 1.4$        & $48.89\pm 1.39$ & $-10.63\pm 7.61$ \\
		ES-MAML    & $6.62\pm 0.97$       & $46.57\pm 1.58$ & $-4.13\pm 8.00$ \\
		IW-IES        & $\bm{21.78\pm 0.92}$    & $\bm{70.95\pm 0.64}$ & $\bm{32.48\pm 2.73}$\\
		\hline
		FS         				& $-17.48\pm 5.68$      & $-8.64\pm 3.51$  		  & $-70.12\pm 8.47$ \\
		CA      				& $15.41\pm 0.96$      & $48.35\pm 1.62$ 	    & $3.57\pm 6.02$   \\
		IW-IES-N          & $17.63\pm 0.89$        & $53.19\pm 1.44$ 	   & $15.31\pm 5.18$ \\
		IW-IES-Qu    	& $21.14\pm 0.94$       & $68.81\pm 0.84$ 		& $29.19\pm 3.08$ \\
		IW-IES-Mix        & $\bm{21.78\pm 0.92}$    & $\bm{70.95\pm 0.64}$ & $\bm{32.48\pm 2.73}$\\
		\cmidrule[\heavyrulewidth]{1-4}
	\end{tabular}
\end{table}

Next, the four variants of IW-IES usually exhibit much faster learning adaptation than FS, especially in the early stage.
The phenomenon indicates that the locomotion skills learned in the original environment can benefit the new learning process a lot.
It also demonstrates the effectiveness of the proposed incremental learning mechanism. 
That is, the previously learned policy empirically performs better than a randomly initialized one because it has learned some of the feature representations of the state-action space.
Further, IW-IES-N, IW-IES-Qu, and IW-IES-Mix always exhibit faster learning adaptation than CA. 
It verifies that using either instance novelty or instance quality as the weighting metric can already enhance the incremental learning performance and enable significantly rapid learning adaptation.
In Mujoco locomotion domains, the weighting metric using instance quality can better boost the incremental learning performance than using instance novelty, and using the mixing variant usually leads to the best learning adaptation to these challenging dynamic environments.
By the instance weighting mechanism that emphasizes new knowledge, IW-IES rapidly guides the policy towards regions of parameter space that better fit in the new environment.
In summary, the results demonstrate that IW-IES is also capable of facilitating learning adaptation to stochastic dynamic environments for these high-dimensional locomotion tasks.

\section{Conclusions}
ES algorithms are recently shown to be capable of solving challenging, high-dimensional RL tasks, while being much faster with many CPUs due to better parallelization. 
In the paper, we investigate incremental ES algorithms for RL in dynamic environments.
We hybridize an instance weighting mechanism with ES to enable rapid learning adaptation, while preserving scalability of ES.
During parameter updating, higher weights are assigned to instances that contain more new knowledge, thus encouraging the search distribution to move towards new promising areas in the parameter space.
The designed weighting metrics, instance novelty and instance quality, ``reinforce" the evolutionary process that searches for new well-behaving policies.
The proposed IW-IES algorithm is tested on traditional navigation and challenging locomotion domains with varying configurations.
Experiments verify that IW-IES is capable of significantly facilitating learning adaptation to various dynamic environments.

Thus, we provide an option for not only taking advantage of the scalability of ES, but also pursuing better learning adaptation to dynamic environments from an incremental learning perspective. 
The latter scenario is supposed to hold for most challenging, real-world domains that ES/RL practitioners will wish to tackle in the future. Our future work will focus on learning adaptation in more challenging dynamic environments where the state-action space changes over time, or learning in more intensively-changing environments (e.g., change between consecutive learning episodes).
Automatic detection and identification of environmental change is also a crucial direction to be addressed.
Another insightful direction would be to conduct empirical investigation on systematically comparing the derivative-free ES algorithms and the derivative-based optimization methods~\citep{Plappert2018parameter} in RL domains, or to develop possible off-policy solutions for incorporating the experience replay mechanism~\citep{schaul2015prioritized} with ES.

\footnotesize 
\bibliography{iwies}
\bibliographystyle{myIEEEtranN}

\begin{IEEEbiography}[{\includegraphics[width=1in,height=1.25in,clip,keepaspectratio]{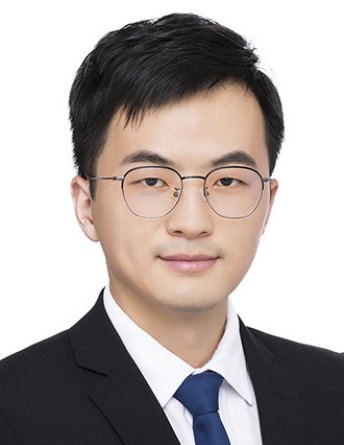}}]{Zhi Wang}
	(Member, IEEE) received the Ph.D. degree in machine learning from the Department of Systems Engineering and Engineering Management, City University of Hong Kong, Hong Kong, China, in 2019, and the B.E. degree in automation from Nanjing University, Nanjing, China, in 2015.
	He is currently an Assistant Professor in the Department of Control and Systems Engineering, Nanjing University.
	He had the visiting position at the University of New South Wales, Canberra, Australia.
	
	His current research interests include reinforcement learning, machine learning, and robotics.
\end{IEEEbiography}

\begin{IEEEbiography}[{\includegraphics[width=1.0in,height=1.25in,clip,keepaspectratio]{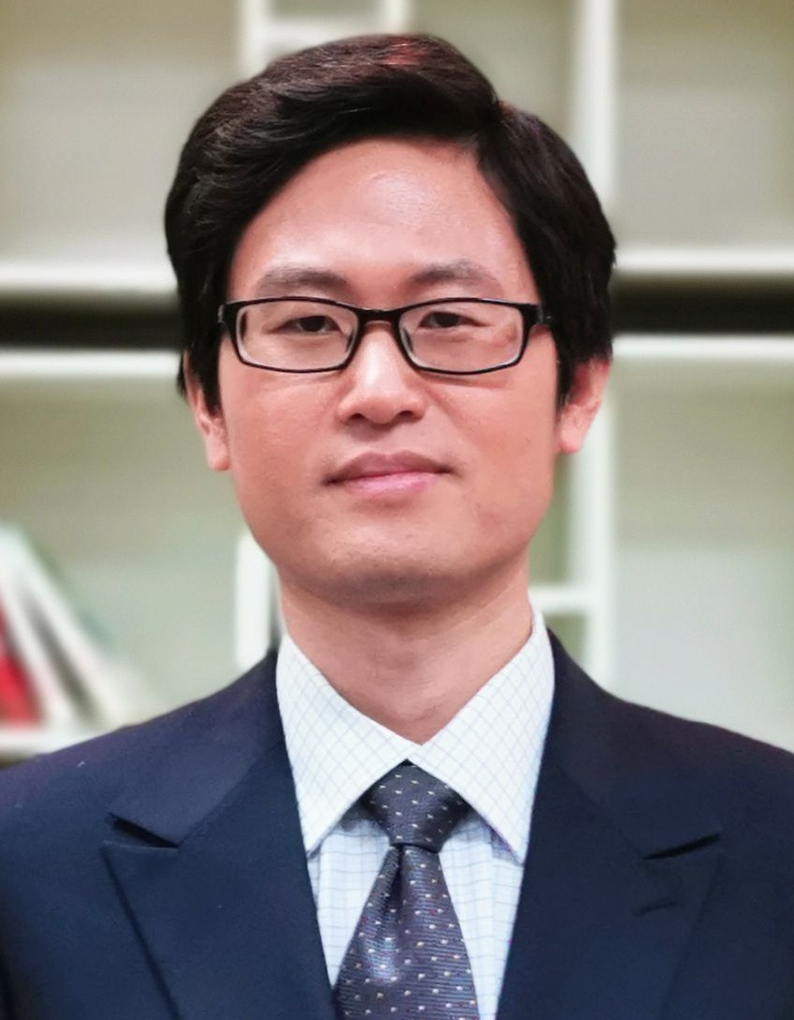}}]{Chunlin Chen}
	(Senior Member, IEEE) received the B.E. degree in automatic control and the Ph.D. degree in control science and engineering from the University of Science and Technology of China, Hefei, China, in 2001 and 2006, respectively.
	
	He was with the Department of Chemistry, Princeton University, Princeton, NJ, USA, from September 2012 to September 2013. He had visiting positions at the University of New South Wales, Canberra, ACT, Australia, and the City University of Hong Kong, Hong Kong. He is currently a Professor and the Head of the Department of Control and Systems Engineering, School of Management and Engineering, Nanjing University, Nanjing, China. His current research interests include machine learning, intelligent control, and quantum control.
	
	Dr. Chen serves as the Chair for the Technical Committee on Quantum Cybernetics, IEEE Systems, Man and Cybernetics Society.
\end{IEEEbiography}

\begin{IEEEbiography}[{\includegraphics[width=1.0in,height=1.25in,clip,keepaspectratio]{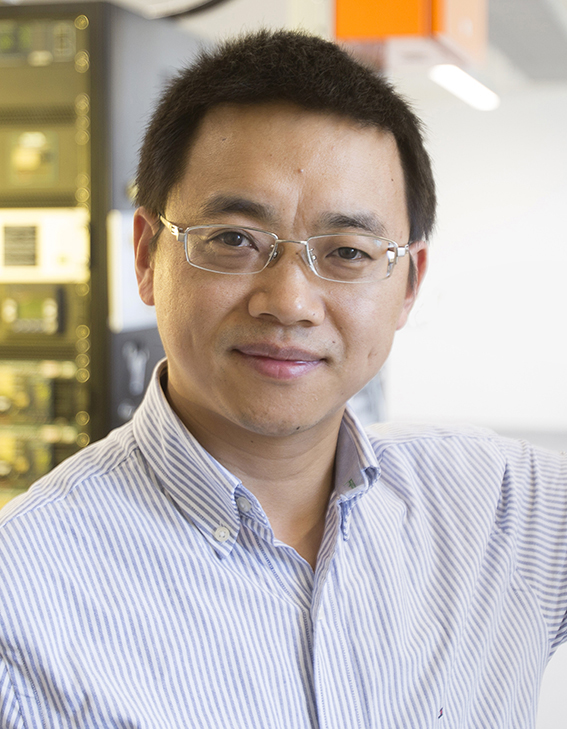}}]{Daoyi Dong}
	received the B.E. degree in automatic control and the Ph.D. degree in engineering from the University of Science and Technology of China, Hefei, China, in 2001 and 2006, respectively.
	
	He was an Alexander von Humboldt Fellow at AKS, University of Duisburg-Essen, Duisburg, Germany.
	He was with the Institute of Systems Science, Chinese Academy of Sciences, Beijing, China, and with Zhejiang University, Hangzhou, China. He had visiting positions at Princeton University, NJ, USA; RIKEN, Wako-Shi, Japan; and The University of Hong Kong, Hong Kong. 
	He is currently a Scientia Associate Professor at the University of New South Wales, Canberra, ACT, Australia. His research interests include quantum control and machine learning.
	
	Dr. Dong was awarded the ACA Temasek Young Educator Award by the Asian Control Association and was a recipient of the International Collaboration Award and the Australian Post-Doctoral Fellowship from the Australian Research Council, and a Humboldt Research Fellowship from the Alexander von Humboldt Foundation of Germany. He is a Member-at-Large, Board of Governors, and the Associate Vice President for Conferences and Meetings, IEEE Systems, Man and Cybernetics Society. He served as an Associate Editor for the IEEE TRANSACTIONSON NEURAL NETWORKS AND LEARNING SYSTEMS from 2015 to 2021. He is currently an Associate Editor of the IEEE TRANSACTIONS ON CYBERNETICS and a Technical Editor of the IEEE/ASME TRANSACTIONS ON MECHATRONICS.
\end{IEEEbiography}

\end{document}